\documentclass[manuscript]{acmart}

\usepackage[utf8]{inputenc} 
\usepackage[T1]{fontenc}    
\usepackage{amsmath, amsthm}
\usepackage{hyperref}       
\usepackage{url}            
\usepackage{booktabs}       
\usepackage{amsfonts}       
\usepackage{nicefrac}       
\usepackage{microtype}      
\usepackage{xcolor}         
\usepackage{threeparttable}
\usepackage{etoolbox}
\usepackage{changepage}
\usepackage{algorithm}
\usepackage{algorithmic}
\usepackage{times}
\usepackage{float}
\usepackage{graphicx}
\usepackage{subfigure}
\usepackage{diagbox}
\usepackage{threeparttable}
\usepackage[version=4]{mhchem}
\usepackage{color, colortbl}
\usepackage{siunitx}
\usepackage{appendix}
\usepackage{booktabs}
\usepackage{multirow}
\usepackage{caption}
\AtBeginDocument{%
  }

\definecolor{LightCyan}{rgb}{0.88,1,1}
\definecolor{LightBlue}{rgb}{0.68, 0.85, 0.90}
\definecolor{LightOrange}{rgb}{1, 0.84, 0.50}
\definecolor{LightPink}{rgb}{0.98, 0.52, 0.9}
\definecolor{LightGreen}{rgb}{0.56, 0.93, 0.56}
\definecolor{bananayellow}{rgb}{1.0, 0.88, 0.21}
\definecolor{capri}{rgb}{0.0, 0.75, 1.0}
\definecolor{corn}{rgb}{0.98, 0.93, 0.36}
\definecolor{applegreen}{rgb}{0.55, 0.71, 0.0}

\begin{document}

\title{Enhanced Agriculture-informed Neural Network by Domain Knowledge}

\author{Ci Lin}
\affiliation{%
	\institution{University of Ottawa}
	\city{Ottawa}
	\country{Canada}
}
\email{clin072@uottawa.ca}
\orcid{0000-0002-8831-6311}

\author{Futong Li}
\affiliation{%
	\institution{University of Ottawa}
	\city{Ottawa}
	\country{Canada}}
\email{fli097@uottawa.ca}
\orcid{0009-0008-5665-7075}

\author{Rose Chong-Wu}
\affiliation{%
	\institution{University of Ottawa}
	\city{Ottawa}
	\country{Canada}}
\email{rchon104@uottawa.ca}
\orcid{0009-0002-5405-9369}

\author{Tet Yeap}
\affiliation{%
	\institution{University of Ottawa}
	\city{Ottawa}
	\country{Canada}}
\email{tyeap@uottawa.ca}
\orcid{0000-0002-6039-6751}

\author{Iluju Kiringa}
\affiliation{%
	\institution{University of Ottawa}
	\city{Ottawa}
	\country{Canada}}
\email{ikiringa@uottawa.ca}
\orcid{0000-0002-9119-9451}

\renewcommand{\shortauthors}{Ci Lin et al.}

\begin{abstract}
	Accurate prediction of nitrous oxide (N$_2$O) emissions from farming is essential for understanding environmental impacts and supporting sustainable farming practices. However, modeling N$_2$O emissions is challenging due to the complex interactions between soil, climate, and biochemical processes, as well as the limited availability of high-quality data. While deep learning models have demonstrated strong predictive capabilities, they often lack interpretability and may fail to generalize under different environmental conditions and farming practice. To address these challenges, Knowledge-enhanced Agriculture-informed Neural Network (KAINN) is proposed by integrating domain knowledge into a hybrid neural–mechanistic modeling framework. Building upon the original Agriculture-informed Neural Network (AINN), the proposed approach explicitly incorporates key environmental processes, including fertilizer diffusion, soil respiration rate (Rh), and water-filled porosity (WFP), to guide the learning process and constrain the solution space.

	We evaluate the proposed framework using CNN-, LSTM-, and Transformer-based architectures across multiple growing seasons and varying input feature configurations. Experimental results demonstrate that KAINN consistently outperforms both purely neural network models and the original AINN in terms of prediction accuracy and generalization, achieving lower Root Mean Square Error (RMSE) and Mean Absolute Error (MAE) and higher coefficient of determination ($R^2$). Furthermore, interface evolution analysis shows that KAINN produces smoother and more physically consistent parameter trajectories with reduced uncertainty, indicating improved interpretability and stability.This work highlights the importance of integrating domain knowledge into deep learning models for environmental applications. The proposed KAINN framework provides a principled and scalable approach for bridging data-driven and mechanistic modeling, enabling more reliable and interpretable prediction of N$_2$O emissions in agricultural systems.
\end{abstract}





\maketitle

\section{Introduction}\label{sec:introduction}

\begin{figure}[htbp]
	\centering   
	\subfigure[N\textsubscript{2}O emissions by country in 2023, measured in CO$_2$-equivalents (GWP100). \protect\cite{jones2024n2o}]{\label{fig:world_n2o_emission}\includegraphics[width=0.33\textwidth]{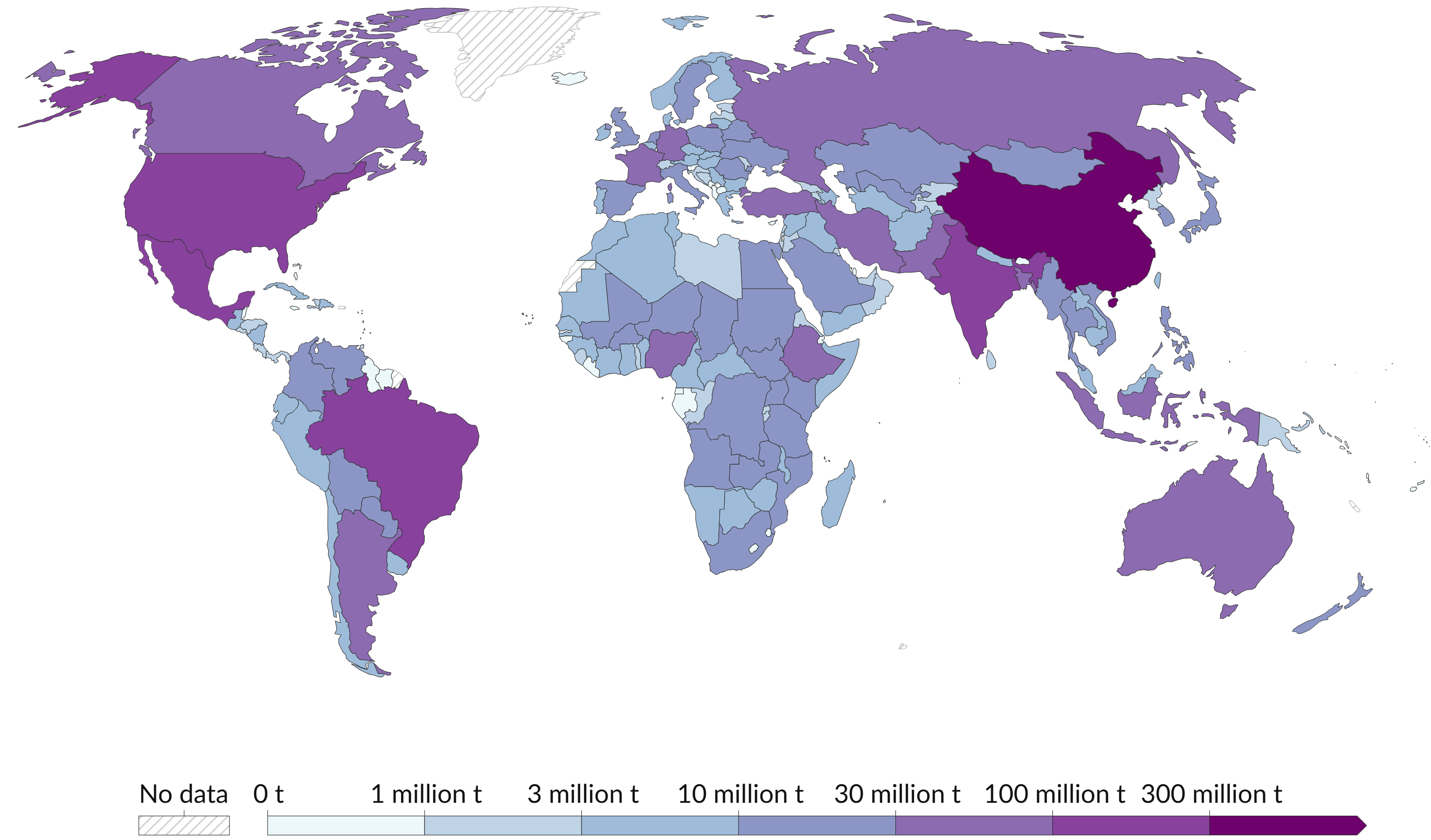}}
	\subfigure[Per capita green house gas emissions (2023). \protect\cite{jones2024ghg}]
	{\label{fig:per_capita}\includegraphics[width=0.33\textwidth]{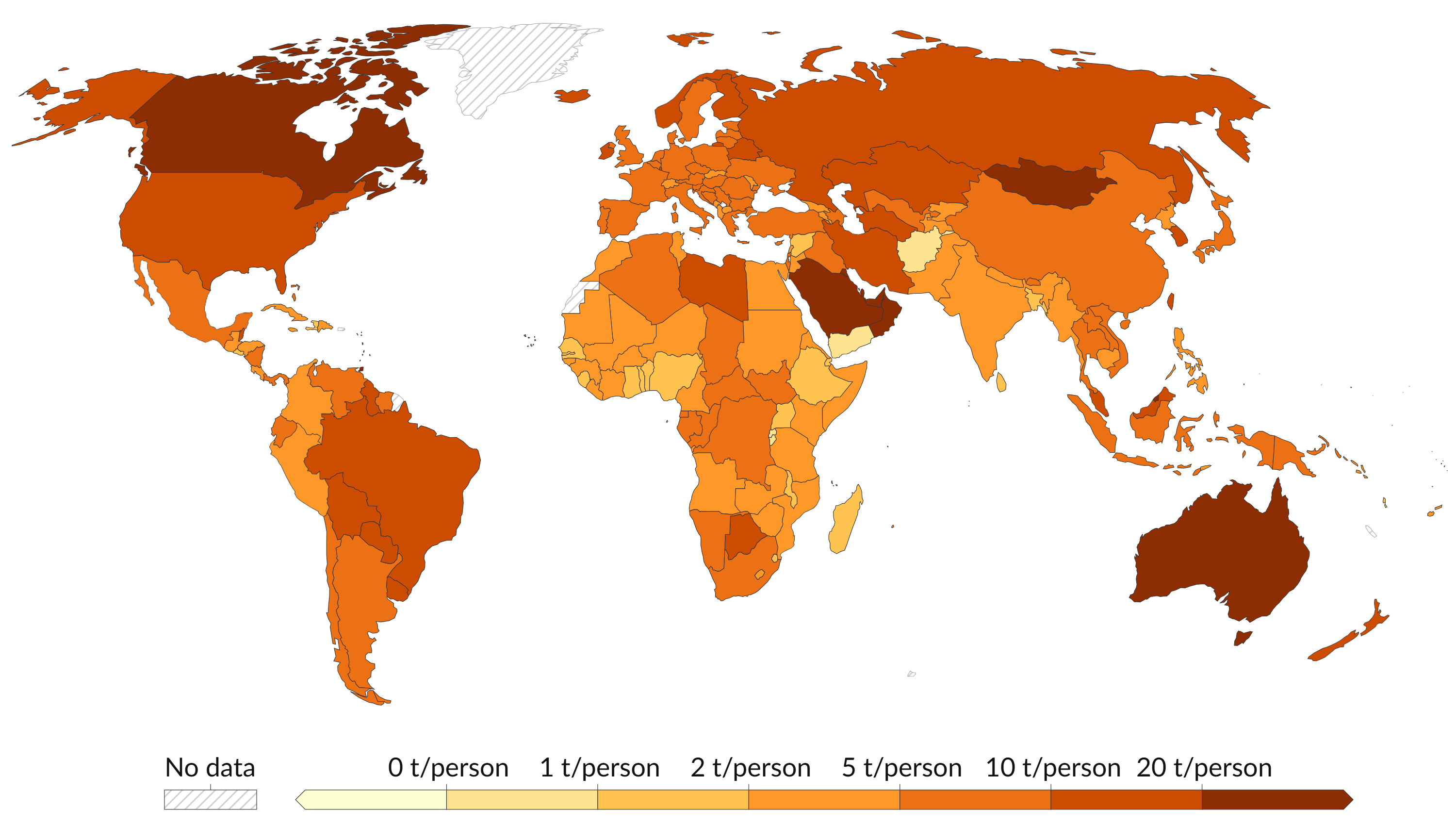}}
	\subfigure[Global N$_2$O Budget (2010--2019). Adapted from the Global Carbon Project, 2024. \protect\cite{tian2024global}]
	{\label{fig:budget_n2o}\includegraphics[width=0.3\textwidth]{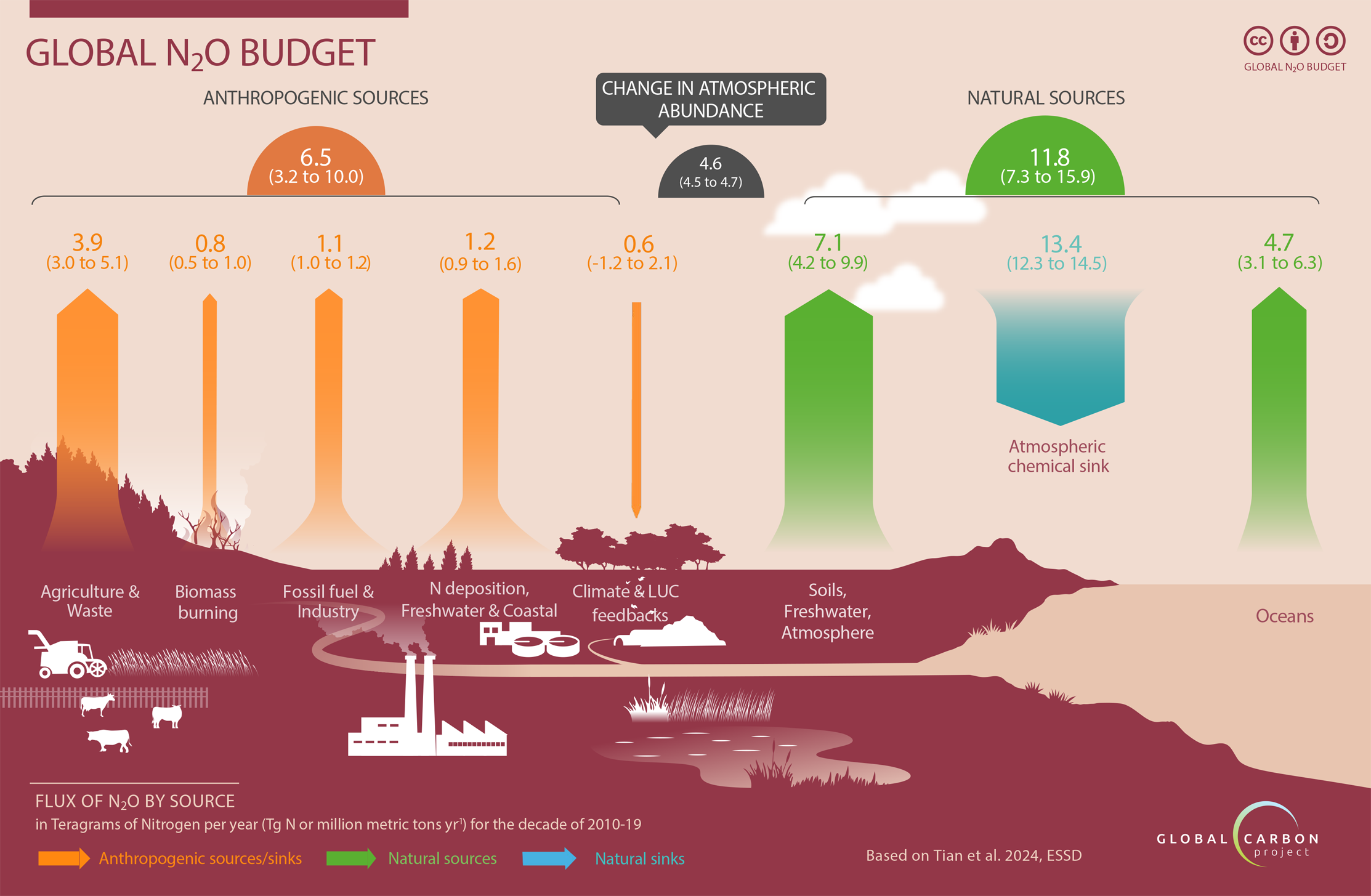}}
	\caption{Global GHG emissions overview.}
	\label{fig:global_overview}
\end{figure}

Climate change remains one of the most urgent global challenges of the 21st century, driven largely by the accumulation of greenhouse gases (GHGs) in the atmosphere \cite{mikhaylov2020global}. Among these gases, nitrous oxide (N$_2$O) has attracted increasing attention due to its exceptionally high global warming potential. According to the Intergovernmental Panel on Climate Change (IPCC), N$_2$O has a global warming potential approximately 273 times greater than that of CO$_2$ over a 100-year time horizon \cite{hartmann2013observations, malik2022drivers, ipcc2021}. In addition to its role in climate warming, N$_2$O is currently the dominant ozone-depleting substance emitted in the 21st century \cite{ravishankara2009nitrous, rapson2014analytical}. A large proportion of anthropogenic N$_2$O emissions originates from agricultural activities, where nitrogen-based fertilizers trigger complex microbial and biochemical processes in the soil \cite{ussiri2013role, cui2014assessing}. As a result, accurately predicting N$_2$O emissions from agricultural systems has become an important task for both climate mitigation and sustainable agricultural management, as indicated in Figure \ref{fig:global_overview}.

However, predicting N$_2$O emissions remains extremely challenging because these emissions arise from highly nonlinear interactions among soil chemistry, microbial dynamics, climate conditions, and agricultural management practices. Early studies relied primarily on empirical or statistical models that extrapolate relationships from observational datasets \cite{freibauer2003controls, leppelt2014nitrous}. Although such models can capture certain large-scale trends, they often fail to represent the complex physical and biological processes that drive N$_2$O emissions, leading to limited predictive capability at fine spatial or temporal scales \cite{butterbach2013nitrous}.

To address these limitations, process-based mechanistic models have been developed to explicitly simulate the underlying carbon and nitrogen cycles in terrestrial ecosystems. Examples include CLM-CN \cite{saikawa2013global}, the Dynamical Land Ecosystem Model (DLEM) \cite{tian2010spatial, xu2017preindustrial}, and LM3V-N \cite{huang2015global}. These models attempt to represent key biogeochemical processes such as nitrification, denitrification, and soil respiration rate (Rh). While mechanistic models provide valuable physical interpretability, they often suffer from large uncertainties in parameterization and struggle to reproduce short-term emission pulses triggered by environmental events such as rainfall or fertilization. Consequently, global N$_2$O emission estimates from natural and agricultural systems vary substantially across different modeling frameworks \cite{ciais2014carbon, tian2012contemporary}.

With the rapid development of machine learning and deep learning techniques, data-driven approaches have increasingly been explored for modeling environmental processes. Various machine learning methods, including multilayer perceptrons (MLPs) \cite{bigaignon2020combination, taki2018comparison}, Random Forests (RF) \cite{saha2021machine}, Long Short Term Memory (LSTM) \cite{Lin2022Stacked}, and Federated Learning (FL) \cite{killeen2025iot}, have been applied to predict N$_2$O emissions using environmental and management data. These models are capable of capturing complex nonlinear relationships and handling high-dimensional multivariate inputs. However, purely data-driven approaches often lack physical interpretability and may produce unreliable predictions when applied to conditions outside the training data distribution.

To combine the strengths of both mechanistic modeling and deep learning, the Agriculture-informed Neural Network (AINN) was recently proposed in 2024 \cite{lin2024agriculture, lin2024analysis, yeapno2024}. AINN integrates deep neural networks with the DLEM, allowing the neural network to learn adjustment factors that interact with the mechanistic processes within the ecosystem model. This hybrid framework demonstrated improved predictive performance compared with purely data-driven models and traditional process-based models. By embedding physical structures into the learning process, AINN represents an important step toward bridging the gap between machine learning and process-based environmental modeling.

\subsection{Motivations and Contributions}

\begin{figure}[htbp]
	\centering
	\includegraphics[width=0.9\linewidth]{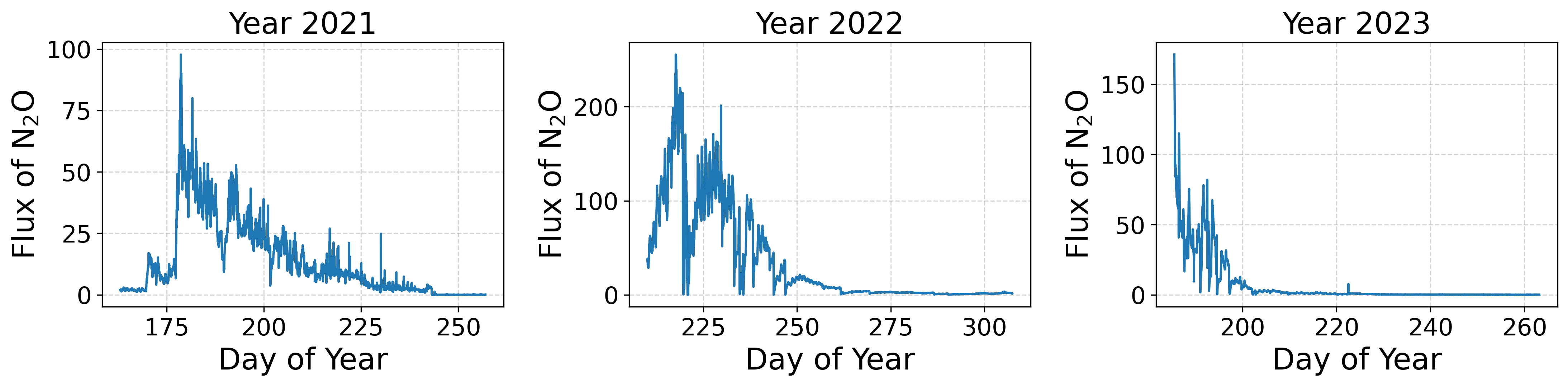}
	\caption{N$_2$O emissions during the growing seasons of 2021, 2022, and 2023, showing distinct patterns influenced by environmental and farming practice. In 2021, a dry year, emissions were delayed until the first major rainfall after fertilizer broadcasting. In 2022, a wet year, emissions occurred immediately following fertilizer application. In 2023, the field was irrigated after fertilizer application, resulting in a higher emission peak and a much shorter emission period. The observed peak emissions were approximately 100 µg N$_2$O-N/m$^2$/sec in 2021, 260 µg N$_2$O-N/m$^2$/sec in 2022, and only 175 µg N$_2$O-N/m$^2$/sec in 2023. These year-to-year differences highlight the complexity of N$_2$O emission processes and the challenges in predicting emissions under varying climatic and management scenarios.}
	\label{fig:main_n2o_emissions_seasons}
\end{figure}

As shown in Figure~\ref{fig:main_n2o_emissions_seasons}, N$_2$O emission patterns across 2021–2023 are presented, highlighting the strong influence of weather conditions, fertilizer management, and irrigation practices on emission dynamics. These observations underscore major challenges for predictive modeling, including data scarcity, high variability, environmental noise, and the need for robust frameworks capable of handling small and uncertain datasets while leveraging domain knowledge for generalization.

In the current AINN framework, the interaction between the neural network and the mechanistic model is primarily implemented through adjustment factors applied to key variables. As a result, several important environmental processes are not explicitly modeled within the neural–mechanistic interface \cite{lin2024agriculture}. For example, processes such as fertilizer diffusion within soil layers, the temperature–moisture dependence of soil respiration (Rh), and the estimation of water-filled porosity (WFP) are not directly incorporated into the learning framework. The absence of these domain-specific mechanisms may limit both the interpretability of the model and its ability to capture critical physical processes governing N$_2$O emissions.

To address these limitations, this paper proposes a knowledge-enhanced Agriculture-informed Neural Network (KAINN) that integrates domain knowledge into AINN. The proposed approach improves the representation of key environmental processes while retaining the flexibility of deep learning models. The main contributions of this work are summarized as follows:

\begin{itemize}
	\item KAINN is proposed, which integrates domain knowledge into the neural–mechanistic interface to improve both prediction accuracy and interpretability for N$_2$O emissions from agricultural systems.
	
	\item Domain knowledge including fertilizer diffusion, soil respiration dynamics, and WFP estimation from soil moisture and temperature are incorporated and systematically evaluated. The temporal evolution of interface parameters provides insights for agricultural practices.
	
	\item A diverse set of variants is developed and evaluated across multiple growing seasons, with training, validation, and testing conducted on agricultural datasets. Their comparative performance analysis helps reveal the underlying mechanisms governing N$_2$O emissions.
\end{itemize}

\subsection{Organization}

The remainder of the paper is organized as follows. Section~\ref{sec:literature_review} reviews existing statistical, process-based, and data-driven models for N$_2$O emission estimation, highlighting their limitations and motivating the need for a hybrid, knowledge-enhanced, and uncertainty-aware approach. Section~\ref{sec:main_methodology} presents the design and theoretical underpinnings of the proposed KAINN framework, an enhanced hybrid model for predicting N$_2$O emissions through the integration of domain knowledge. Section~\ref{sec:experiment_discussion} describes the experimental setup, data preprocessing, and model evaluation. Performance comparisons are conducted across LSTM-, CNN-, and Transformer-based architectures, with and without knowledge integration, using RMSE, MAE, and $R^2$ as evaluation metrics. Section~\ref{sec:conclusion} summarizes the contributions of KAINN, emphasizing its improved predictive performance, interpretability, and robustness under data-scarce and noisy environmental conditions. The limitations of the framework are also discussed, along with potential directions for future research in N$_2$O emission modeling.

\section{Literature Reviews}\label{sec:literature_review}
Process-based modeling is essential in evaluating and predicting changes to the Nitrogen cycle and N$_2$O emissions in response to various global factors. Several process-based models have been applied to estimate N$_2$O emissions from natural and agricultural soils over different time and space scales.

As a rain event-driven process-based model, the DeNitrification-DeComposition (DNDC) has been widely used to model Nitrogen cycling in croplands and grasslands under various management practices \cite{li1992model, congreves2016predicting, cui2014assessing, giltrap2010dndc}. Furthermore, considering the simultaneous occurrence of oxic (an environment or condition where oxygen is present) and anoxic (an environment or condition where oxygen is absent or present at deficient levels) sites for nitrification and denitrification processes via the concept of a dynamic anaerobic balloon, the PnET-N-DNDC was developed for forest areas. This model considers explicitly the effect of freezing and thawing on soil moisture \cite{kiese2005regional, li2000process}; however, its application is limited since it needs detailed growth information from the field \cite{miehle2006assessing}. Based on the ``hole in pipe model'' concept, the Carnegie-Ames-Stanford approach (CASA) Biosphere model was developed to estimate the N$_2$O flux from natural soil \cite{firestone1989microbiological, potter1993terrestrial, potter1996process}. However, this model's nitrification and denitrification modules are primarily conceptually driven, regardless of the level of detail of microbial processes \cite{firestone1989microbiological}. Although the Daily Century (DAYCENT) model \cite{parton1996generalized} is capable of simulating N$_2$O emission from cropland \cite{del2002simulated, del2009global} and pasture \cite{abdalla2010testing}, it does not include the process of oxygen diffusion and consumption \cite{butterbach2004temporal}. The ECOSYS \cite{grant2001modeling} could provide a reliable estimation of N$_2$O flux from soils at any temporal or spatial scale via three-dimensional flux equations; however, it needs a relatively large amount of data and is difficult to parameterize. Therefore, it could not be widely adopted in agriculture \cite{chen2008n}.

The Farm ASSEssment Tool (FASSET) model can accurately predict year-round N$_2$O emissions for European countries; instead of utilizing the equation of denitrification, it uses semi-empirical functions \cite{chatskikh2005simulation, olesen2002comparison}. Based on the Community Earth System Model (CLM), Carbon and Nitrogen cycle version 3.5, the CLM-CN 3.5 model was developed to estimate the emission of N$_2$O. This model simulates the seasonality and interannual variability of global N$_2$O emissions while it is only well suited to capture the variability of emissions from specific ecosystems, such as forests, and significant uncertainties are caused by its coarse simulation resolution \cite{saikawa2013global}. Derived from the Integrated Biosphere Simulator (IBIS), TRIPLEX-GHG couples nitrification and denitrification processes to quantify N$_2$O emissions from natural forests and grasslands \cite{zhu2014modelling, zhang2017process}. Although TRIPLEX-GHG is appropriate to simulate N$_2$O emissions from different forest and grassland land types under varying environmental conditions on a global scale, it is less robust in modeling N$_2$O uptake and peaks during periods of snowmelt. LM3V-N could accurately estimate the mean values of N$_2$O flux; however, it still has site-to-site and temporal mismatches \cite{huang2015global}. Vegetation Integrative Simulator for Trace gases (VISIT) is an integrated model for simulating biogeochemical interactions (carbon and nitrogen cycles). It was developed to estimate the atmosphere–ecosystem exchanges of greenhouse gases (CO$_2$, CH$_4$, and N$_2$O) and to determine the global warming potential (GWP), taking into account the radiative forcing effect of each gas \cite{inatomi2010greenhouse, ito2012use}. The DLEM has been successfully applied to simulate N$_2$O flux from terrestrial ecosystems throughout North America and globally. However, the daily time step of this model may underestimate N$_2$O emissions because the possibility of high pulses was not considered \cite{tian2010spatial, tian2013global}.

The estimations of global terrestrial N$_2$O budgets and spatiotemporal patterns have varied significantly among these models. These discrepancies are primarily due to differences in the input data used, the structure of the models, and the methods used to determine model parameters. Despite being calibrated to specific locations, process-based biogeochemical models typically cannot accurately forecast daily or monthly emissions with more than a \SI{20}{\%} margin of error. These challenges have prompted a shift toward data-driven approaches. Machine learning models such as Multilayer Perceptrons (MLPs) \cite{bigaignon2020combination, taki2018comparison}, RF \cite{saha2021machine}, and LSTM \cite{Lin2022Stacked} offer improved empirical performance, especially in data-scarce conditions. However, these methods often struggle to model multivariate time series effectively and lack physical interpretability. To overcome these shortcomings, the AINN was proposed \cite{lin2024agriculture, killeen2025iot}, integrating a deep learning framework with the DLEM to embed chemical insights into the learning process. Despite its advances, the original AINN still faces challenges in robustness, adaptability, and uncertainty estimation under diverse farming conditions, motivating the development of improved hybrid frameworks.

\section{Methodology}\label{sec:main_methodology}

\subsection{A Brief Introduction to Dynamic Land Ecosystem Model}\label{sec:dlem_component}

The DLEM is a comprehensive, terrestrial, process-based model that integrates biogeochemical cycles, hydrological processes, and vegetation dynamics to simulate the structural and functional dynamics of land ecosystems. It accounts for a wide range of factors, including climate, atmospheric compositions (such as CO$_2$ concentration and nitrogen deposition), land cover changes, and land management practices (such as harvesting, crop rotation, and fertilization) \cite{tian2012century}.

The DLEM framework is organized around five core components: biophysics, plant physiology, soil biogeochemistry, dynamic vegetation, and land use and management. In this study, we investigate N$_2$O emissions, which are influenced by the interactions among all five components.

\subsubsection{Emission of N$_2$O}

The emission of N$_2$O from soils is primarily dominated by biological nitrification and denitrification processes, as shown in Equation \ref{equ:n2o_sum}. Both nitrification and denitrification are functions of soil temperature, WFP, soil moisture, soil respiration, and so on.

\begin{equation}\label{equ:n2o_sum}
	\begin{aligned}
		& G_{N_2O} = (N_{nit} + N_{denit}) \times f(T_{soil}) \times (1 - f(wfp)) \\
		& f(T_{soil}) = \frac{1}{1 + e^{-0.64 + 0.08T_{soil}}} \\
		& f(wfp) = 0.0116 + \frac{1.36}{1 + e^{-\frac{wfp-0.815}{0.0896}}} \\
	\end{aligned}
\end{equation}

\noindent
where $G_{N_2O}$ is the N$_2$O produced from soil, $N_{nit}$ is the daily nitrification rate (\si{g N m^{-2} day^{-1}}), $N_{denit}$ is the daily denitrification rate (\si{g N m^{-2} day^{-1}}), $f (T_{soil})$ is the scaled factor of soil temperature ($T_{soil}$, \si{^\circ C}) on the N$_2$O emission process (unitless), and $f(wfp)$ is the effects of WFP.

\subsubsection{Nitrification}
Nitrification is a biological process that involves the conversion of ammonia (\ce{NH_{3}}) and ammonium (\ce{NH_{4}^{+}}) into nitrite (\ce{NO_{2}^{-}}) and nitrate (\ce{NO_{3}^{-}}) by soil microorganisms, which are nitrifying bacteria. This process is essential to the nitrogen cycle and is crucial in maintaining soil fertility and ecosystem productivity. The nitrification rate is calculated as follows:

\begin{equation}\label{equ:n2o_nit}
	\begin{aligned}
		& N_{nit} = k_{nit} \times f(T_{Nsoil}) \times g(wfp) \times D_{NH_4} \\
		& f(T_{Nsoil}) = 7.24 \times e^{-3.432 + 0.618 \times T_{soil} \times (1 - 0.5 \times T_{soil}/36.9)} \\
		& g(wfp) = - 12.904 \times wfp^4 + 17.651 \times wfp^3 + \\
		& \quad \quad \quad 5.5368 \times wfp^2 + 0.9975 \times wfp - 0.0243
	\end{aligned}
\end{equation}

\noindent
where $k_{nit}$ is the daily maximum fraction of \ce{NH_{4}^{+}} that is converted into \ce{NO_{3}^{-}} and nitrogen gases, $f(T_{Nsoil})$ is the soil temperature's effect on nitrification (unitless), and $g(wfp)$ represents the effect of WFP on the nitrification process.

\subsubsection{Denitrification}
Denitrification is the biological process that converts \ce{NO_{3}^{-}} into nitric oxide (NO), N$_2$O, and dinitrogen (N$_2$). The denitrification rate are calculated as follows:

\begin{equation}\label{equ:n2o_denit}
	\begin{aligned}
		& N_{denit} = N_{pot_{denit}} \times f(T_{soil}) \times f(wfp) \times f(D_{NO_3}) \\
		& N_{pot_{denit}} = (0.151 + 0.015 \times P_{clay}) \times Rh \times k_{den} \\
		& f(D_{NO_3}) = 1.17 \times \frac{D_{NO_3}}{32.7 + D_{NO_3}} \\
		& D_{NO_3} = av_{NO_3}/ BD_{soil}
	\end{aligned}
\end{equation}

\noindent
where $N_{pot_{denit}}$ is the potential rate of denitrification (\si{g N m^{-2} day^{-1}}), $P_{clay}$ is the percentage of clay content in soil, $Rh$ is the soil respiration rate (\si{g C m^{-2} day^{-1}}), $k_{den}$ is a parameter depending on plant functional type to tune the potential denitrification rate (\si{g N m^{-2}day^{-1}}), $f (D_{NO_3})$ represents the effect of \ce{NO_{3}^{-}} concentration (\si{g N g^{-1}}), $av_{NO_3}$ is the \ce{NO_{3}^{-}} content in the soil per unit area (\si{g N m^{-2}}), and $BD_{soil}$ is the soil bulk density (\si{g m^{-3}}).

\subsection{A Knowledge-Enhanced AINN Framework}

\begin{table}[htbp]
	\centering
	\begin{threeparttable}
		\caption{Parameter ranges used in the original AINN for data transformation.}
		\begin{tabular}
			{m{2cm}m{1cm}m{1cm}m{1cm}m{1cm}m{1cm}m{1cm}m{1cm}m{1cm}}
			\toprule $[1]$ & D$_{NO_3}$ & D$_{NH_4}$ & Temp & WFP & Rh  & P$_{clay}$ & k$_{den}$ & k$_{nit}$ \\
			\midrule maximum & 250 & 250 & 40 & 1 & 100 & 1 & 1 & 1 \\
			\midrule minimum & 0 & 0 & 10 & 0 & 0 & 0 & 0 & 0 \\
			\bottomrule
		\end{tabular}
		\label{table:para_range}
		\begin{tablenotes}
			\footnotesize
			\item[1] The units of D$_{NO_3}$ and D$_{NH_4}$ are \si{g\ N\ g^{-1}\ soil}. Temperature is measured in \si{^\circ C}. WFP and P$_{clay}$ are expressed as percentages (\si{\%}). Rh is measured in \si{g\ C\ m^{-2}\ day^{-1}}. k$_{den}$ is measured in \si{g\ N\ m^{-2}\ day^{-1}}, and k$_{nit}$ is unitless.
		\end{tablenotes}
	\end{threeparttable}
\end{table}

\begin{figure}[htbp]
	\centering
	\subfigure[Proposed KAINN]{
		\includegraphics[width=0.45\textwidth]{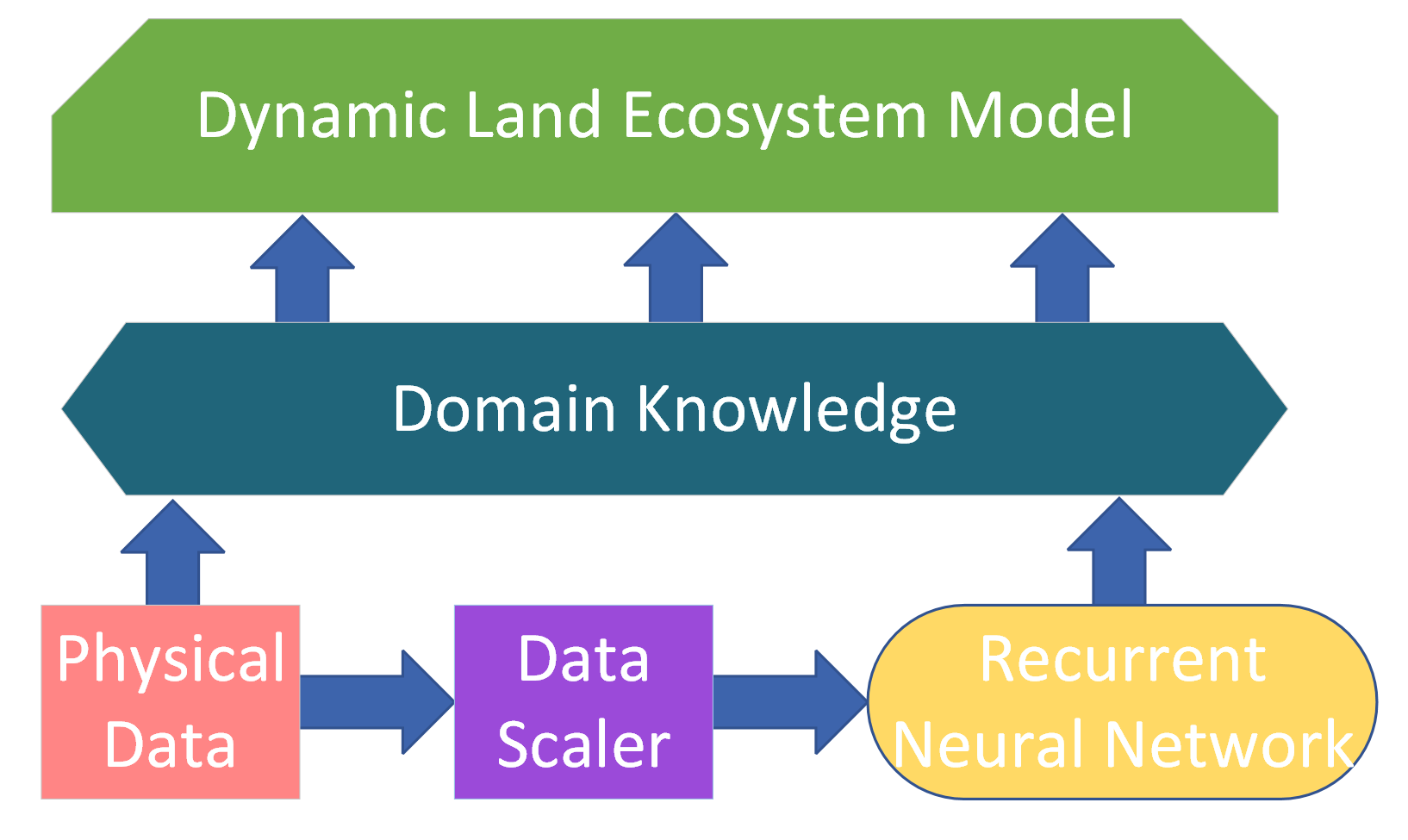}
		\label{fig:kainn}
	}
	\subfigure[Original AINN]{
		\includegraphics[width=0.45\textwidth]{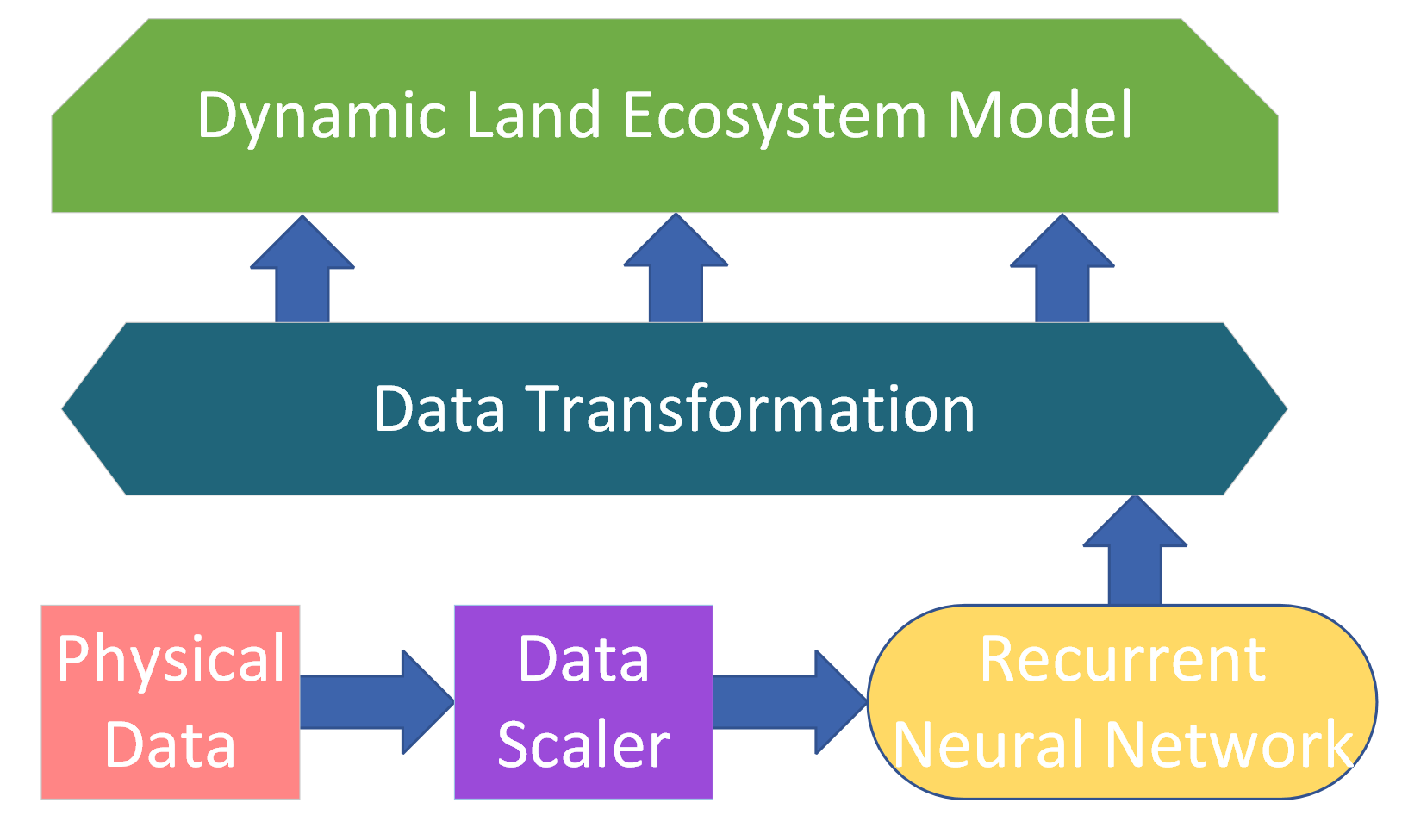}
		\label{fig:main_oainn}
	}
	\caption{Architectural comparison between the original AINN and the proposed KAINN. The original AINN integrates an RNN with the structural framework of the DLEM, where the neural network outputs are scaled and transformed into physical variables before being passed to the DLEM. In contrast, KAINN incorporates domain knowledge (denoted as ``K'') to explicitly model environmental processes and agronomic mechanisms. These knowledge-guided variables are used to compute key quantities such as Rh, WFP, and NO$_3^{-}$ concentration, which are then provided to the DLEM to predict N$_2$O emissions. This tighter integration reduces information entropy and improves both generalization and interpretability.}
	\label{fig:main_ainn_architecture}
\end{figure}

To provide a high-level understanding of the proposed framework, we illustrate the overall design of KAINN and AINN in Figure~\ref{fig:main_ainn_architecture}. In the original AINN framework, the RNN component outputs an 8-dimensional vector at each time step. These outputs correspond to intermediate variables that are mapped to key parameters of the DLEM, including soil temperature (T$_{soil}$), WFP, nitrification rate (k$_{nit}$), denitrification potential (k$_{den}$), soil clay fraction (P$_{clay}$), Rh, and nitrogen-related concentrations such as NO$_3^{-}$ and NH$_4^{+}$. 

Since the neural network outputs are constrained within the range $[0,1]$, they are subsequently transformed into physically meaningful values using predefined parameter ranges, as summarized in Table~\ref{table:para_range}. The transformation is defined in Equation \ref{equ:transformation_layer}.
\begin{equation}\label{equ:transformation_layer}
	P = C \cdot (P_{\max} - P_{\min}) + P_{\min}.
\end{equation}
In this formulation, the neural network learns a latent representation that indirectly controls the physical parameters, while the DLEM enforces mechanistic consistency.

In contrast, the proposed KAINN framework introduces a more explicit and physically grounded integration of domain knowledge. Although the neural network outputs remain within the range $[0,1]$, they are no longer directly mapped to physical parameters through simple scaling. Instead, they are combined with observed physical variables, such as soil moisture and soil temperature, to estimate the parameters required by the DLEM. This process incorporates domain knowledge into the transformation pipeline, allowing the model to compute physically meaningful quantities through structured relationships rather than purely data-driven mappings.

As a result, KAINN provides a more faithful representation of environmental processes by aligning the learned representations with physically interpretable variables. The detailed formulation of the domain knowledge and the corresponding transformation process are presented in Section~\ref{subsec:domain_knowledge}.

\subsubsection{A Case Study of AINN and KAINN}

\begin{table}[htbp]
	\centering
	\begin{threeparttable}
	\caption{Architecture of CNN-DLEM.}
	\begin{tabular}{lccc}
		\toprule
		Layer\tnote{a} & Input Size & Output Size & Parameters \\
		\midrule
		Conv1D  & $[B,T,6]$   & $[B,T,100]$ & 1,900 \\
		Conv1D  & $[B,T,100]$ & $[B,T,100]$ & 30,100 \\
		Linear  & $[B,T,100]$ & $[B,T,100]$ & 10,000 \\
		Linear  & $[B,T,100]$ & $[B,T,8]$   & 800 \\
		DLEM 	& $[B,T,8]$   & $[B,T]$     & -- \\
		\midrule
		Total & -- & -- & 42,800 \\
		\bottomrule
	\end{tabular}
	\label{tab:cnn_dlem_arch}
	\begin{tablenotes}
		\footnotesize
		\item[a] For convolutional layers, the input tensor is internally reshaped from $[B,T,D]$ to $[B,D,T]$ and then restored. These permutation operations do not introduce trainable parameters and are omitted.
	\end{tablenotes}
	\end{threeparttable}
\end{table}

\begin{table}[htbp]
	\centering
	\begin{threeparttable}
		\caption{Architecture of CNN-KDLEM.}
		\begin{tabular}{lccc}
			\toprule
			Layer\tnote{a} & Input Size & Output Size & Parameters \\
			\midrule
			Conv1D\tnote{b}  & $[B,T,3]$ & $[B,T,100]$ & 1,000 \\
			Conv1D  & $[B,T,100]$ & $[B,T,100]$ & 30,100 \\
			Linear  & $[B,T,100]$ & $[B,T,100]$ & 10,000 \\
			Linear  & $[B,T,100]$ & $[B,T,8]$ & 800 \\
			KDLEM\tnote{c} & $[B,T,8]$ + Physical Data & $[B,T]$ & -- \\
			Aggregation\tnote{d} & $\sum_{i=1}^{2} [B,T]$ & $[B,T]$ & 0 \\
			\midrule
			Total & -- & -- & 41,900 \\
			\bottomrule
		\end{tabular}
		\label{tab:cnn_kdlem_arch}
		\begin{tablenotes}
			\footnotesize
			\item[a] For convolutional layers, the input tensor is internally reshaped from $[B,T,D]$ to $[B,D,T]$ and then restored. These permutation operations do not introduce trainable parameters and are omitted.
			\item[b] In this layer, the input data is split into two parts, each containing 3 features that represent measurements collected from a specific soil depth $i$. Therefore, compared with Table~\ref{tab:cnn_dlem_arch}, the input channels in CNN-KDLEM are 3 rather than 6.
			\item[c] Before feeding the neural network output into the KDLEM module, the corresponding physical variables are combined with the learned latent states to predict N$_2$O emissions.
			\item[d] Since we compute the N$_2$O emissions at different depths individually using the same model, these values are eventually summed to obtain the final N$_2$O emission.
		\end{tablenotes}
	\end{threeparttable}
\end{table}

To help readers better understand the improvement of KAINN over AINN, we take CNN-DLEM and CNN-KDLEM as representative examples to illustrate the key differences between these two frameworks. We assume that the input data are organized into $T$ time steps with six features, which are used to predict N$_2$O emissions from farming.

Table~\ref{tab:cnn_dlem_arch} presents the architecture of CNN-DLEM, which follows the standard AINN paradigm. In this model, all input features are concatenated and processed jointly through a single convolutional feature extractor. The extracted temporal features are then mapped to an 8-dimensional latent state at each time step, which serves as the input to the DLEM module. The DLEM component subsequently enforces mechanistic constraints to produce the final prediction. This design treats the soil system as a unified entity, where all depth-related information is implicitly mixed within a single latent representation.

In contrast, Table~\ref{tab:cnn_kdlem_arch} illustrates the architecture of CNN-KDLEM, which reflects the core idea of KAINN. Instead of processing all features jointly, the input is decomposed into multiple depth-specific groups, where each group corresponds to measurements collected from a particular soil layer. Each depth-specific input is independently processed by a shared CNN feature extractor to generate a corresponding latent state. These latent states are then combined with depth-specific physical variables and passed to the KDLEM module, which applies the same underlying physical mechanisms across all depths. Finally, the outputs from different depths are aggregated to produce the overall emission prediction.

The fundamental difference between these two models lies in how domain knowledge is integrated into the neural network. CNN-DLEM relies on a single global representation and applies mechanistic constraints at the aggregated level, whereas CNN-KDLEM adopts a structured, depth-aware design that explicitly models the heterogeneity of soil processes. By decomposing the input into depth-specific components while maintaining shared physical laws, CNN-KDLEM is able to capture both local variations and global consistency.

This design offers two major advantages. First, it improves physical interpretability by aligning each branch with a specific soil depth and corresponding environmental processes. Second, it enhances predictive performance by allowing the model to learn depth-dependent patterns that are otherwise obscured in a fully aggregated representation. As a result, CNN-KDLEM offers a more faithful representation of real-world soil dynamics than CNN-DLEM.

\subsection{Integration of Domain Knowledge into DLEM Using Environmental and Biochemical Signals}\label{subsec:domain_knowledge}

DLEM is a process-based framework that simulates N$_2$O emissions by capturing key interactions among environmental and biological processes. The model primarily focuses on two emission pathways—nitrification and denitrification—which are governed by eight critical parameters: soil temperature, WFP, soil moisture, Rh, clay content, \ce{NH$_4^+$} concentration, \ce{NO$_3^-$} concentration, and soil bulk density. Accurate estimation of these parameters is crucial to ensure reliable emission predictions and robust generalization across varying field conditions.

To improve the prediction accuracy and generalization capability of DLEM within our hybrid modeling framework, we introduce the following domain-informed modifications:

\begin{itemize}
	\item Modeling NH$_4^+$ and NO$_3^-$ concentrations according to the Fick’s Second Law.
	\item Replacing estimated soil temperature with observed temperature measurements.
	\item Estimating WFP from observed soil moisture data.
	\item Estimating the Rh from the soil temperature and soil moisture.
\end{itemize}

From an information-theoretic perspective, these modifications can be interpreted as reducing the entropy of the model’s hypothesis space. Let \( X \) represent the set of input features (e.g., soil temperature, moisture, NH$_4^+$, NO$_3^-$) and \( Y \) denote the target variable representing N$_2$O emissions. In a standard data-driven setting, the model must learn the mapping \( X \rightarrow Y \) under high uncertainty due to falsified architecture, noisy, incomplete, or poorly estimated features, leading to higher conditional entropy \( H(Y \mid X) \).

By embedding domain knowledge—such as chemistry-informed fertilizer transformation curves and field-observed soil temperature—we constrain the variability of input features and remove non-informative noise. This reduction in input uncertainty effectively lowers \( H(X) \), which in turn reduces \( H(Y \mid X) \), making the predictive distribution more concentrated and reliable.

Moreover, the mutual information between inputs and outputs, defined as \(I(X; Y) = H(Y) - H(Y \mid X),\) is increased through the integration of meaningful, mechanistically grounded signals. A higher \( I(X; Y) \) implies that the model’s inputs are more informative about the outputs, thereby enhancing both learning efficiency and generalization capability. Thus, these improvements to DLEM not only embed environmental understanding but also serve to optimize the information flow within the hybrid framework.

\subsubsection{Estimating $\ce{NH_4^+}$ and $\ce{NO_3^-}$ concentrations using Diffusion Model}

Fertilizer transport in soil is primarily governed by diffusion, which describes the spontaneous movement of dissolved solutes from regions of higher concentration to regions of lower concentration due to concentration gradients. In the soil environment, nutrients such as nitrogen move through water-filled pores even when there is no bulk water flow. Unlike advection, which requires water movement, diffusion occurs under still conditions and therefore plays a critical role in nutrient redistribution after fertilizer application.

To describe this process, the vertical movement of fertilizer in the soil profile is modeled using a one-dimensional diffusion equation:

\begin{equation}
	\frac{\partial C(z,t)}{\partial t}
	=
	\frac{\partial}{\partial z}
	\left(
	D(\theta(z,t),T(z,t))
	\frac{\partial C(z,t)}{\partial z}
	\right)
\end{equation}

\noindent
where $C(z,t)$ denotes the solute concentration at soil depth $z$ and time $t$, $z$ represents soil depth (cm), $t$ denotes time (day), and $D(\theta,T)$ is the effective diffusion coefficient \cite{hillel2003introduction}.

The diffusion coefficient is assumed to depend on soil moisture and soil temperature, reflecting the influence of environmental conditions on nutrient mobility. The diffusivity is modeled using an empirical relationship:

\begin{equation}
	D(\theta,T)
	=
	D_0
	\left(\frac{\theta}{100}\right)^n
	\left(\frac{T}{T_{\mathrm{ref}}}\right)^m
\end{equation}

\noindent
where $\theta$ denotes volumetric soil moisture (\%), $T$ is soil temperature (K), $D_0$ is a reference diffusion coefficient, $T_{\mathrm{ref}}$ is a reference temperature, and $n$ and $m$ are empirical parameters controlling the sensitivity of diffusion to soil moisture and temperature, respectively. This formulation allows the diffusion rate to dynamically vary with soil environmental conditions, enabling the simulation to capture how fertilizer spreads through the soil profile under changing moisture and temperature conditions \cite{moldrup2001tortuosity}.

While the diffusion model described above governs the vertical transport of fertilizer within the soil profile, the chemical forms of nitrogen also evolve over time due to microbial processes. In particular, urea (CON$_2$H$_4$) applied as fertilizer is first hydrolyzed to ammonium (NH$_4^+$), which is subsequently oxidized into nitrate (NO$_3^-$) through nitrification. To capture this temporal transformation process, we model NH$_4^+$ as an exponentially decreasing function and NO$_3^-$ as an exponentially increasing function over time after fertilizer application. The concentrations of NO$_3^-$ and NH$_4^+$ are therefore expressed as functions of DSFA, as shown in Equation~\ref{equ:estimated_fertilizer}.

\begin{equation}\label{equ:estimated_fertilizer}
	\begin{aligned}
		& D_{NO_3^-} \propto \exp\left(\alpha \cdot \frac{DSFA}{DSFA_{\text{MAX}}}\right) \times \exp \left( \beta \left(1 -  \cdot \frac{DSFA}{DSFA_{\text{MAX}}} \right) \right) \\
		& D_{NH_4^+} \propto \exp \left(  \alpha \left(1 -  \cdot \frac{DSFA}{DSFA_{\text{MAX}}} \right) \right)
	\end{aligned}
\end{equation}

\noindent
where \( DSFA_{\text{MAX}} \) denotes the time span from the initial fertilizer application to the point at which N$_2$O emissions have diminished to near zero, and \( \alpha \) and \( \beta \) are empirical coefficients that indicate the rate of change. Typically, when \( \alpha \) or \( \beta = 1, 2, 3 \), KAINN achieves relatively better performance. These values are then scaled to the range \([0, 1]\) and combined with the adjustment factors generated by the deep learning model to estimate the real-time concentrations of NO$_3^-$ and NH$_4^+$. For NO$_3^-$, the first term represents contributions from other chemical processes. In particular, in this context, NH$_4^+$ is converted to NO$_3^-$, while NO$_3^-$ is further transformed into N$_2$O and other compounds, which is presented by the second term. Typically, an overshoot coefficient is also applied to the estimated values to allow for a certain level of tolerance and variability.

\subsubsection{Using Observed Temperature Measurements}

Soil temperature plays a critical role in regulating microbial activity, especially in processes such as nitrification and denitrification, which are primary pathways for N$_2$O emissions. 

To improve the fidelity of the model, we replace the internally estimated soil temperature with field-measured temperature data. This substitution ensures that temperature-dependent response functions—such as \( f(T_{\text{soil}}) \) and \( f(T_{\text{Nsoil}}) \) in Equations~\ref{equ:n2o_sum} and~\ref{equ:n2o_nit}—are directly influenced by real-time environmental observations. As a result, the model can more precisely capture the dynamics of microbial transformation rates and reduce the uncertainty associated with thermal response mismatches.

\subsubsection{Estimating WFP from Soil Moisture}

WFP describes the fraction of soil pore volume occupied by water and is commonly used to characterize soil moisture conditions influencing microbial activity and nitrogen transformations. WFPS can be estimated from volumetric soil moisture by normalizing it with soil porosity:

\begin{equation}
WFP = \frac{\theta}{\phi},
\end{equation}

\noindent
where $\theta$ represents the volumetric soil moisture (cm$^3$ cm$^{-3}$) and $\phi$ denotes the soil porosity. Soil porosity can be estimated from soil bulk density using

\begin{equation}
\phi = 1 - \frac{\rho_b}{\rho_p},
\end{equation}

\noindent
where $\rho_b$ is the soil bulk density and $\rho_p$ is the particle density. In Area X.O, it is typically  sandy soils, and bulk density is generally higher and porosity is lower compared with finer-textured soils, which influences the resulting WFP values and soil aeration conditions. The WFP values estimated from soil moisture measurements are therefore used to represent soil moisture conditions affecting nitrogen transformation processes in the model \cite{fu2024effects}.

Combined the analysis discussed above, the WFP is estimated empirically from observed soil moisture (SM), as shown in Equation \ref{equ:estimated_wfp}.

\begin{equation}\label{equ:estimated_wfp}
	WFP = \frac{\beta \times \text{SM}}{100}
\end{equation}

\noindent
where, \( \beta \) represents soil porosity, which is passed from deep learning model. This approach allows the model to respond more accurately to hydrological variations, which is essential for simulating soil oxygen availability and microbial activity.

\subsubsection{Modeling Soil Respiration as a Function of Soil Temperature and Soil Moisture}

Soil respiration is a critical process driven by soil microbial activity and root metabolism, playing a vital role in regulating ecosystem carbon cycling. Among the most influential environmental drivers of soil respiration are soil temperature and soil moisture.

Soil Rh increases exponentially with temperature, as microbial and enzymatic activities accelerate in warmer conditions. This behavior is consistent with thermodynamic principles and is typically modeled using an exponential function. The response of soil Rh to soil moisture is more complex. While increased moisture can enhance microbial processes by supporting substrate diffusion and enzyme mobility, Rh decreases beyond a certain moisture threshold due to oxygen limitation. This relationship is often represented as a Gaussian-like curve, peaking at an optimal moisture level.

Soil moisture \(SM\) is assumed to be within the range \([0, 1]\) based on its definition. Soil temperature \(ST\) is scaled by a constant \(T_{\text{max}}\), the expected maximum soil temperature, so that \(ST / T_{\text{max}} \in [0, 1]\).

Based on the relationship between the Rh, soil temperature, and soil moisture, we define the Equation \ref{equ:soil_respiration}.

\begin{equation}\label{equ:soil_respiration}
	R = \exp\left( k \cdot \frac{ST}{T_{\text{max}}} \right) \cdot \exp\left( -\frac{(SM - SM_{\text{opt}})^2}{2\sigma^2} \right) 
\end{equation}

\noindent
where $k$ is the temperature sensitivity coefficient, $T_{\text{max}}$ is the maximum temperature tolerable by soil bacteria, $SM_{\text{opt}}$ is the optimal soil moisture, and $\sigma$ controls the spread of the moisture response \cite{lloyd1994temperature, yan2018moisture}.

Equation \ref{equ:soil_respiration} captures the exponential growth of Rh with temperature and the unimodal response to moisture. The resulting \(R\) will be normalized to retain only the structural relationship. It will then be multiplied by a base respiration rate, which is set manually according to different environmental conditions and agricultural practices. Through this manner, we decouple the relationship and the specified value that will be set in the real scenario.

For the clay content percentage (P$_{clay}$), denitrification rate (k$_{den}$), and nitrification rate (k$_{nit}$) in the DLEM, it is normally in the range of [0, 1], therefore, it is supplied by the deep learning component in the AINN model.

\section{Experiments and Analysis} \label{sec:experiment_discussion}

\subsection{Data Preprocessing}

\begin{figure}[htbp]
	\centering
	\includegraphics[width=0.8\linewidth]{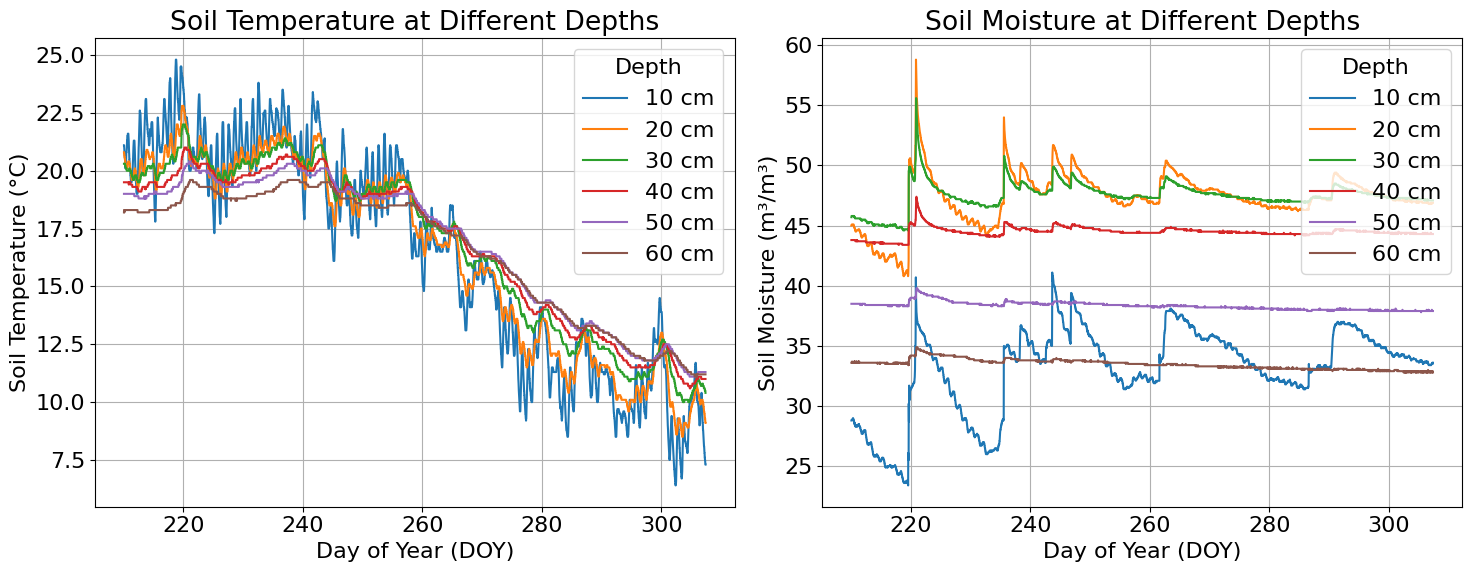}
	\caption{Temporal profile of soil temperature (left) and soil moisture (right) measured at depths of 10, 20, 30, 40, 50, and 60 cm over the 2022 growing season (x-axis: day of year). The figure shows that shallow soil layers respond more rapidly and exhibit greater variability due to environmental fluctuations, whereas deeper layers display more stable and gradual changes. This contrast in temporal behavior across soil depths highlights the advantage of fusing multi-depth measurements to reduce surface-level noise and enhance the model’s generalization performance.}
	\label{fig:sm_st}
\end{figure}

The raw environmental data collected from the field exhibit varying numerical ranges depending on the type of measurement. In order to feed these inputs into a deep learning model, data scaling is necessary, since neural networks generally perform best when features are scaled to the range \([0, 1]\), accelerating convergence and ensures numerical stability during training.

However, in practice, it is often not possible to know the true minimum and maximum values of each stochastic variable across all possible datasets, especially for unseen testing data. To ensure fairness and consistency across both training and inference phases, we standardize the input data using biologically reasonable fixed ranges. Specifically, we define a scaling window that covers the typical conditions in which microbial activity is most likely to occur, i.e., the sweet zone for soil biochemical reactions. 

Consequently, we set the scaling range for soil temperature to \([10, 30]\) and for soil moisture to \([10, 60]\), regardless of the exact data distribution in individual samples. This approach ensures that the scaling process is biologically grounded, model-agnostic, and robust to unexpected data values during testing. The full set of variable ranges used in our preprocessing pipeline is shown in Table~\ref{table:min_max_parameter}.

\begin{table}[htbp]
	\centering
	\begin{threeparttable}
		\caption{Data Ranges Used for Scaling in the Proposed Approach}
		\begin{tabular}{m{2.5cm} m{1.5cm} m{1.5cm} m{1.5cm} m{1.5cm}}
			\toprule
			$[1]$ & FN$_2$O & DSFA & Temperature & Moisture \\
			\midrule
			Maximum & 260 & 90 & 30 & 60 \\
			Minimum & 0 & 0 & 10 & 10 \\
			\bottomrule
		\end{tabular}
		\label{table:min_max_parameter}
		\begin{tablenotes}
			\footnotesize
			\item[1] Units: N$_2$O flux is in \si{ppb}, DSFA is day since the first application, soil moisture is in \si{\%}, and soil temperatures is in \si{^\circ C}.
		\end{tablenotes}
	\end{threeparttable}
\end{table}

The Equation \ref{equ:min_max_scaler} is used for scaling data.

\begin{equation}\label{equ:min_max_scaler}
	X^{'} = \frac{X - X_{min}}{X_{max} - X_{min}}
\end{equation}

\noindent
where $X^{'}$ is the transformed value, which is normally in the range of [0, 1], and $X$ is the original data. $X_{max}$ and $X_{min}$ are the maximum and minimum values shown in Table \ref{table:min_max_parameter}. The transformed data $X^{'}$ could be represented by a vector at one time step, as shown in the following:

\begin{equation}\nonumber
	\begin{bmatrix}
		\mathbf{DSFA}_{t} & \mathbf{ST}_{t} & \mathbf{SM}_{t}
	\end{bmatrix}
\end{equation}

\noindent
where \textbf{DSFA}, \textbf{SM}$_{t}$, and \textbf{ST}$_{t}$ represent the day since the first application, estimated soil moisture, and estimated soil temperature, respectively, at time step $t$. 

To predict the emission of N$_2$O at time step $t$, information from the previous $n$ time steps is required. Therefore, the input data can be characterized as a data matrix, as shown below:
\begin{equation}\nonumber
	\begin{bmatrix} 
		\mathbf{DSFA}_{t-n:t-1}^T & \mathbf{ST}_{t-n:t-1}^T & \mathbf{SM}_{t-n:t-1}^T 
	\end{bmatrix}
\end{equation}
\noindent
where \textbf{DSFA}$_{t-n:t-1}$, \textbf{SM}$_{t-n:t-1}$, \textbf{ST}$_{t-n:t-1}$ are row vectors consisting of the values of \textbf{DSFA}, \textbf{SM}, and \textbf{ST} from time step $t-n$ to $t-1$. $T$ is the transpose operation.

\subsection{Evaluation Metrics}

\begin{table}[htbp]
	\centering
	\small
	\caption{Evaluation metrics used to assess the performance of the proposed model.}
	\begin{tabular}{m{1.5cm}m{4cm}m{7cm}}
		\toprule
		Metric & Formula & Description \\
		\midrule
		RMSE &
		$\displaystyle
		RMSE = \sqrt{\frac{1}{n}\sum_{i=1}^{n}(y_i-\hat{y}_i)^2}
		$ &
		Root Mean Squared Error measures the average magnitude of prediction errors, giving higher weight to larger errors. \\
		\midrule
		MAE &
		$\displaystyle
		MAE = \frac{1}{n}\sum_{i=1}^{n}|y_i-\hat{y}_i|
		$ &
		Mean Absolute Error measures the average absolute difference between predicted and observed values. \\
		\midrule
		$R^2$ &
		$\displaystyle
		R^2 = 1 - \frac{\sum_{i=1}^{n}(y_i-\hat{y}_i)^2}{\sum_{i=1}^{n}(y_i-\bar{y})^2}
		$ &
		Coefficient of determination measuring the proportion of variance explained by the model. \\
		\bottomrule
	\end{tabular}
	\label{tab:evaluation_metrics}
\end{table}

As shown in Table \ref{tab:evaluation_metrics}, to comprehensively evaluate the performance of the proposed approach, we employ a set of metrics that primarily measure prediction accuracy and goodness-of-fit for the predicted N$_2$O emissions. The prediction accuracy is assessed using the Root Mean Squared Error (RMSE), Mean Absolute Error (MAE), and the coefficient of determination ($R^2$). RMSE measures the average magnitude of prediction errors while penalizing larger deviations more strongly due to the squared error term. MAE evaluates the average absolute difference between predicted and observed values, providing a more interpretable measure of typical prediction error. The $R^2$ metric measures the proportion of variance in the observed data that is explained by the model and provides an overall indication of goodness-of-fit.

\subsection{Experimental Setup}

To evaluate the proposed KAINN framework, we develop a variety of models based on three major deep learning architectures: LSTM, CNN, and Transformer. For each architecture, six variants are constructed, differing in the integration of domain knowledge (\textbf{K}) with DLEM and their combinations.

All models are trained on data from the 2021 growing season, validated on the 2022 growing season, and tested on the 2023 growing season, ensuring a realistic evaluation of generalization across different years. The models are trained using the Adam optimizer with a fixed learning rate of $1 \times 10^{-4}$, and RMSE is used as the loss function. To prevent overfitting, early stopping is applied based on validation loss thresholds. Each configuration is trained ten times to mitigate stochastic variations and ensure stable performance evaluation.

\subsection{Performance Comparison of LSTM-, CNN-, and Transformer-Based Models}\label{subsec:nn_ainn}

\subsubsection{Performance Analysis with Three Features}
\begin{table}
\centering
\small
\setlength{\tabcolsep}{3pt}
\caption{Performance comparison of various models across validation (2022) and testing (2023) datasets with 3 features. \\ Best results within each (year, step) group are highlighted in bold.}
\begin{tabular}{l c c c c | c c c}
	\toprule
	& & \multicolumn{3}{c}{2022} & \multicolumn{3}{c}{2023} \\
	\cmidrule(lr){3-5} \cmidrule(lr){6-8}
	Variant & Step & RMSE & MAE & R$^2$ & RMSE & MAE & R$^2$ \\
	\midrule
	CNN & 12 & $36.098 \pm 3.936$ & $24.835 \pm 4.883$ & $0.529 \pm 0.105$ & $21.019 \pm 4.489$ & $17.052 \pm 5.039$ & $-0.533 \pm 0.666$ \\
	CNN-DLEM & 12 & $33.388 \pm 12.730$ & $20.769 \pm 8.633$ & $0.549 \pm 0.451$ & $11.307 \pm 2.697$ & $6.220 \pm 0.951$ & $0.552 \pm 0.262$ \\
	CNN-KDLEM & 12 & $\mathbf{28.306 \pm 0.569}$ & $\mathbf{16.429 \pm 0.574}$ & $\mathbf{0.713 \pm 0.011}$ & $\mathbf{10.563 \pm 0.243}$ & $\mathbf{4.848 \pm 0.456}$ & $\mathbf{0.628 \pm 0.017}$ \\
	\midrule
	CNN & 24 & $35.932 \pm 3.702$ & $24.353 \pm 3.851$ & $0.533 \pm 0.102$ & $19.304 \pm 5.755$ & $15.331 \pm 5.732$ & $-0.342 \pm 0.863$ \\
	CNN-DLEM & 24 & $33.427 \pm 12.716$ & $20.864 \pm 8.625$ & $0.548 \pm 0.450$ & $11.647 \pm 2.675$ & $6.559 \pm 1.736$ & $0.526 \pm 0.259$ \\
	CNN-KDLEM & 24 & $\mathbf{28.113 \pm 0.430}$ & $\mathbf{16.796 \pm 0.299}$ & $\mathbf{0.717 \pm 0.009}$ & $\mathbf{10.550 \pm 0.400}$ & $\mathbf{5.227 \pm 0.801}$ & $\mathbf{0.628 \pm 0.029}$ \\
	\midrule
	CNN & 48 & $36.945 \pm 3.831$ & $25.234 \pm 4.214$ & $0.507 \pm 0.101$ & $21.661 \pm 5.553$ & $17.311 \pm 5.480$ & $-0.657 \pm 0.794$ \\
	CNN-DLEM & 48 & $29.229 \pm 0.695$ & $18.989 \pm 0.653$ & $0.694 \pm 0.015$ & $10.758 \pm 0.484$ & $6.376 \pm 1.453$ & $0.614 \pm 0.035$ \\
	CNN-KDLEM & 48 & $\mathbf{28.327 \pm 0.314}$ & $\mathbf{16.692 \pm 0.223}$ & $\mathbf{0.713 \pm 0.006}$ & $\mathbf{10.562 \pm 0.326}$ & $\mathbf{4.834 \pm 0.555}$ & $\mathbf{0.628 \pm 0.023}$ \\
	\midrule
	\midrule
	LSTM & 12 & $36.801 \pm 5.858$ & $25.523 \pm 6.012$ & $0.504 \pm 0.175$ & $21.604 \pm 8.241$ & $17.804 \pm 8.286$ & $-0.760 \pm 1.493$ \\
	LSTM-DLEM & 12 & $28.137 \pm 0.555$ & $17.187 \pm 0.724$ & $0.717 \pm 0.011$ & $\mathbf{10.833 \pm 0.321}$ & $5.670 \pm 0.655$ & $\mathbf{0.608 \pm 0.023}$ \\
	LSTM-KDLEM & 12 & $\mathbf{27.724 \pm 0.428}$ & $\mathbf{16.446 \pm 0.634}$ & $\mathbf{0.725 \pm 0.009}$ & $11.048 \pm 0.517$ & $\mathbf{4.893 \pm 0.628}$ & $0.592 \pm 0.039$ \\
	\midrule
	LSTM & 24 & $34.968 \pm 5.011$ & $23.938 \pm 5.186$ & $0.554 \pm 0.137$ & $20.064 \pm 7.778$ & $15.999 \pm 8.013$ & $-0.523 \pm 1.261$ \\
	LSTM-DLEM & 24 & $27.915 \pm 0.952$ & $17.725 \pm 1.008$ & $0.721 \pm 0.019$ & $11.108 \pm 0.851$ & $6.931 \pm 1.316$ & $0.586 \pm 0.065$ \\
	LSTM-KDLEM & 24 & $\mathbf{27.459 \pm 0.437}$ & $\mathbf{16.698 \pm 0.582}$ & $\mathbf{0.730 \pm 0.009}$ & $\mathbf{10.858 \pm 0.397}$ & $\mathbf{5.304 \pm 0.878}$ & $\mathbf{0.606 \pm 0.029}$ \\
	\midrule
	LSTM & 48 & $35.363 \pm 6.393$ & $24.754 \pm 5.930$ & $0.539 \pm 0.196$ & $20.425 \pm 6.731$ & $16.508 \pm 6.986$ & $-0.527 \pm 1.259$ \\
	LSTM-DLEM & 48 & $27.805 \pm 0.810$ & $18.263 \pm 0.803$ & $0.723 \pm 0.016$ & $13.046 \pm 1.668$ & $9.096 \pm 1.440$ & $0.424 \pm 0.154$ \\
	LSTM-KDLEM & 48 & $\mathbf{27.043 \pm 0.514}$ & $\mathbf{16.189 \pm 0.712}$ & $\mathbf{0.738 \pm 0.010}$ & $\mathbf{10.739 \pm 0.250}$ & $\mathbf{5.499 \pm 0.566}$ & $\mathbf{0.615 \pm 0.018}$ \\
	\midrule
	\midrule
	Transformer & 12 & $31.273 \pm 3.217$ & $21.165 \pm 3.762$ & $0.647 \pm 0.074$ & $15.701 \pm 3.314$ & $11.346 \pm 4.287$ & $0.145 \pm 0.357$ \\
	Transformer-DLEM & 12 & $\mathbf{27.374 \pm 0.591}$ & $17.313 \pm 0.931$ & $\mathbf{0.732 \pm 0.011}$ & $12.244 \pm 0.823$ & $6.985 \pm 1.318$ & $0.498 \pm 0.069$ \\
	Transformer-KDLEM & 12 & $27.813 \pm 0.477$ & $\mathbf{16.722 \pm 0.378}$ & $0.723 \pm 0.009$ & $\mathbf{11.672 \pm 0.301}$ & $\mathbf{6.936 \pm 0.490}$ & $\mathbf{0.546 \pm 0.023}$ \\
	\midrule
	Transformer & 24 & $32.263 \pm 6.403$ & $22.458 \pm 5.776$ & $0.614 \pm 0.177$ & $16.221 \pm 5.075$ & $12.534 \pm 5.295$ & $0.046 \pm 0.667$ \\
	Transformer-DLEM & 24 & $27.293 \pm 0.821$ & $17.409 \pm 0.746$ & $0.733 \pm 0.016$ & $\mathbf{11.435 \pm 0.266}$ & $\mathbf{6.931 \pm 0.769}$ & $\mathbf{0.564 \pm 0.020}$ \\
	Transformer-KDLEM & 24 & $\mathbf{26.867 \pm 0.754}$ & $\mathbf{17.057 \pm 0.453}$ & $\mathbf{0.741 \pm 0.014}$ & $12.566 \pm 0.930$ & $8.281 \pm 0.997$ & $0.471 \pm 0.080$ \\
	\midrule
	Transformer & 48 & $30.454 \pm 3.014$ & $20.434 \pm 2.951$ & $0.665 \pm 0.066$ & $14.473 \pm 2.507$ & $10.859 \pm 3.013$ & $0.283 \pm 0.260$ \\
	Transformer-DLEM & 48 & $28.255 \pm 0.944$ & $18.171 \pm 0.731$ & $0.714 \pm 0.019$ & $\mathbf{11.858 \pm 0.591}$ & $\mathbf{8.043 \pm 0.951}$ & $\mathbf{0.530 \pm 0.047}$ \\
	Transformer-KDLEM & 48 & $\mathbf{27.224 \pm 0.737}$ & $\mathbf{17.105 \pm 0.628}$ & $\mathbf{0.735 \pm 0.014}$ & $13.257 \pm 0.678$ & $8.754 \pm 0.480$ & $0.413 \pm 0.059$ \\
	\midrule
	\bottomrule
\end{tabular}
\label{tab:ainn_3_features_block}
\end{table}

Table~\ref{tab:ainn_3_features_block} presents a comparison of various models on the validation (2022) and testing (2023) datasets using three input features. Although the three features supplied limited information for the model to predict the emission of N$_2$O accurately, the results still provide several useful observations.

First, the proposed KAINN generally demonstrates improved performance over the original AINN across most configurations, particularly in terms of RMSE, MAE and R$^2$. For example, in CNN-based models, CNN-KDLEM consistently achieves lower prediction errors across all temporal resolutions. Similar trends can also be observed in LSTM-based models, where LSTM-KDLEM leads to modest but consistent improvements, especially on the 2023 dataset. These results suggest that incorporating additional domain knowledge can enhance predictive accuracy.

Second, the advantage of KAINN becomes more apparent under different environmental conditions and farming practice(from 2022 to 2023). Although all models experience some performance degradation on the testing dataset, KAINN tends to maintain more stable RMSE and $R^2$ values. Compared with AINN, KAINN also shows improved stability in several cases, suggesting that the refined knowledge integration may contribute to better robustness.

Third, the results indicate that purely data-driven models may be insufficient when only limited features are available. Across all architectures, the baseline CNN, LSTM, and Transformer models exhibit higher prediction errors and less stable performance compared with AINN and KAINN. This observation highlights the importance of incorporating domain knowledge when modeling complex environmental processes.

Finally, the impact of temporal window size appears to be relatively moderate compared with the effect of knowledge integration. KAINN maintains consistent performance across different sequence lengths, suggesting that incorporating meaningful inductive bias may be more influential than increasing temporal context alone.

\subsubsection{Performance Analysis with Six Features}

\begin{table}[htbp]
	\centering
	\small
	\setlength{\tabcolsep}{3pt}
	\caption{Performance comparison of various models across validation (2022) and testing (2023) datasets with 6 features. \\ Best results within each (year, step) group are highlighted in bold.}
	\begin{tabular}{l c c c c | c c c}
		\toprule
		& & \multicolumn{3}{c}{2022} & \multicolumn{3}{c}{2023} \\
		\cmidrule(lr){3-5} \cmidrule(lr){6-8}
		Variant & Step & RMSE & MAE & R$^2$ & RMSE & MAE & R$^2$ \\
		\midrule
		CNN & 12 & $40.181 \pm 7.309$ & $26.708 \pm 5.575$ & $0.405 \pm 0.219$ & $22.607 \pm 6.190$ & $18.044 \pm 5.870$ & $-0.819 \pm 1.039$ \\
		CNN-DLEM & 12 & $34.715 \pm 12.300$ & $20.984 \pm 8.562$ & $0.520 \pm 0.441$ & $11.127 \pm 2.756$ & $5.184 \pm 0.873$ & $0.565 \pm 0.266$ \\
		CNN-KDLEM & 12 & $\mathbf{27.949 \pm 0.286}$ & $\mathbf{16.440 \pm 0.346}$ & $\mathbf{0.720 \pm 0.006}$ & $\mathbf{10.799 \pm 0.386}$ & $\mathbf{4.606 \pm 0.466}$ & $\mathbf{0.611 \pm 0.028}$ \\
		\midrule
		CNN & 24 & $40.850 \pm 5.489$ & $27.706 \pm 4.823$ & $0.393 \pm 0.160$ & $22.497 \pm 4.840$ & $17.984 \pm 4.596$ & $-0.757 \pm 0.689$ \\
		CNN-DLEM & 24 & $30.901 \pm 0.566$ & $18.296 \pm 0.700$ & $0.658 \pm 0.013$ & $10.698 \pm 1.065$ & $5.687 \pm 1.667$ & $0.615 \pm 0.078$ \\
		CNN-KDLEM & 24 & $\mathbf{28.068 \pm 0.484}$ & $\mathbf{16.614 \pm 0.289}$ & $\mathbf{0.718 \pm 0.010}$ & $\mathbf{10.475 \pm 0.204}$ & $\mathbf{4.995 \pm 0.476}$ & $\mathbf{0.634 \pm 0.014}$ \\
		\midrule
		CNN & 48 & $37.903 \pm 4.619$ & $26.435 \pm 4.523$ & $0.479 \pm 0.128$ & $21.169 \pm 3.584$ & $17.222 \pm 3.528$ & $-0.532 \pm 0.516$ \\
		CNN-DLEM & 48 & $32.300 \pm 1.219$ & $19.770 \pm 0.873$ & $0.626 \pm 0.028$ & $10.663 \pm 0.657$ & $5.396 \pm 0.794$ & $0.620 \pm 0.046$ \\
		CNN-KDLEM & 48 & $\mathbf{28.276 \pm 0.594}$ & $\mathbf{16.569 \pm 0.409}$ & $\mathbf{0.714 \pm 0.012}$ & $\mathbf{10.513 \pm 0.065}$ & $\mathbf{5.013 \pm 0.581}$ & $\mathbf{0.632 \pm 0.005}$ \\
		\midrule
		\midrule
		LSTM & 12 & $42.929 \pm 10.214$ & $29.613 \pm 7.858$ & $0.307 \pm 0.343$ & $23.235 \pm 8.421$ & $18.785 \pm 7.598$ & $-1.012 \pm 1.438$ \\
		LSTM-DLEM & 12 & $30.546 \pm 1.474$ & $19.462 \pm 1.723$ & $0.665 \pm 0.032$ & $11.248 \pm 0.779$ & $6.492 \pm 1.497$ & $0.576 \pm 0.059$ \\
		LSTM-KDLEM & 12 & $\mathbf{27.893 \pm 0.715}$ & $\mathbf{17.015 \pm 0.737}$ & $\mathbf{0.721 \pm 0.014}$ & $\mathbf{10.903 \pm 0.441}$ & $\mathbf{5.673 \pm 0.942}$ & $\mathbf{0.603 \pm 0.033}$ \\
		\midrule
		LSTM & 24 & $38.326 \pm 6.415$ & $26.077 \pm 5.939$ & $0.461 \pm 0.185$ & $20.992 \pm 6.410$ & $16.740 \pm 6.598$ & $-0.592 \pm 1.008$ \\
		LSTM-DLEM & 24 & $30.634 \pm 0.801$ & $19.183 \pm 1.240$ & $0.664 \pm 0.018$ & $11.581 \pm 0.724$ & $7.827 \pm 1.219$ & $0.551 \pm 0.056$ \\
		LSTM-KDLEM & 24 & $\mathbf{27.438 \pm 0.655}$ & $\mathbf{16.600 \pm 0.806}$ & $\mathbf{0.730 \pm 0.013}$ & $\mathbf{11.191 \pm 0.471}$ & $\mathbf{5.922 \pm 1.116}$ & $\mathbf{0.582 \pm 0.035}$ \\
		\midrule
		LSTM & 48 & $39.226 \pm 4.160$ & $26.797 \pm 4.414$ & $0.444 \pm 0.121$ & $21.441 \pm 4.265$ & $17.148 \pm 4.189$ & $-0.587 \pm 0.660$ \\
		LSTM-DLEM & 48 & $31.141 \pm 1.163$ & $20.036 \pm 1.401$ & $0.652 \pm 0.026$ & $13.015 \pm 1.359$ & $9.094 \pm 1.759$ & $0.430 \pm 0.118$ \\
		LSTM-KDLEM & 48 & $\mathbf{26.958 \pm 0.743}$ & $\mathbf{16.406 \pm 0.741}$ & $\mathbf{0.740 \pm 0.014}$ & $\mathbf{10.718 \pm 0.451}$ & $\mathbf{5.755 \pm 0.832}$ & $\mathbf{0.616 \pm 0.032}$ \\
		\midrule
		\midrule
		Transformer & 12 & $33.960 \pm 4.645$ & $23.411 \pm 4.277$ & $0.580 \pm 0.118$ & $17.189 \pm 5.324$ & $13.626 \pm 5.297$ & $-0.070 \pm 0.665$ \\
		Transformer-DLEM & 12 & $30.126 \pm 0.983$ & $19.534 \pm 1.402$ & $0.675 \pm 0.021$ & $12.057 \pm 0.996$ & $8.103 \pm 1.489$ & $0.512 \pm 0.081$ \\
		Transformer-KDLEM & 12 & $\mathbf{27.560 \pm 0.901}$ & $\mathbf{16.671 \pm 0.535}$ & $\mathbf{0.728 \pm 0.018}$ & $\mathbf{12.004 \pm 0.466}$ & $\mathbf{7.512 \pm 0.831}$ & $\mathbf{0.519 \pm 0.038}$ \\
		\midrule
		Transformer & 24 & $32.562 \pm 1.965$ & $23.182 \pm 1.986$ & $0.619 \pm 0.046$ & $16.019 \pm 2.309$ & $13.033 \pm 2.157$ & $0.129 \pm 0.268$ \\
		Transformer-DLEM & 24 & $29.756 \pm 0.903$ & $19.151 \pm 1.287$ & $0.683 \pm 0.019$ & $\mathbf{11.693 \pm 0.907}$ & $7.861 \pm 1.572$ & $\mathbf{0.542 \pm 0.071}$ \\
		Transformer-KDLEM & 24 & $\mathbf{26.369 \pm 0.769}$ & $\mathbf{16.775 \pm 0.503}$ & $\mathbf{0.751 \pm 0.015}$ & $12.197 \pm 0.574$ & $\mathbf{7.794 \pm 0.740}$ & $0.503 \pm 0.046$ \\
		\midrule
		Transformer & 48 & $32.865 \pm 5.808$ & $22.852 \pm 5.538$ & $0.603 \pm 0.162$ & $15.995 \pm 4.291$ & $12.826 \pm 4.423$ & $0.092 \pm 0.596$ \\
		Transformer-DLEM & 48 & $28.493 \pm 0.615$ & $18.336 \pm 0.898$ & $0.709 \pm 0.012$ & $\mathbf{11.648 \pm 0.792}$ & $\mathbf{7.543 \pm 1.003}$ & $\mathbf{0.546 \pm 0.060}$ \\
		Transformer-KDLEM & 48 & $\mathbf{27.300 \pm 0.923}$ & $\mathbf{16.909 \pm 0.653}$ & $\mathbf{0.733 \pm 0.018}$ & $13.372 \pm 1.258$ & $9.079 \pm 0.716$ & $0.399 \pm 0.119$ \\
		\midrule
		\bottomrule
	\end{tabular}
	\label{tab:ainn_6_features_block}
\end{table}

Table~\ref{tab:ainn_6_features_block} presents the performance of CNN-, LSTM-, and Transformer-based models using six input features across the validation (2022) and testing (2023) datasets. Several key observations can be made.

Similar to the results obtained with three input features, KAINN consistently improves predictive performance over AINN and purely data-driven models in both CNN- and LSTM-based architectures. In the 2022 dataset, KAINN achieves lower RMSE and MAE, along with higher $R^2$, across all temporal resolutions. This trend remains stable in the 2023 dataset, where KAINN maintains lower prediction errors and improved generalization. These results indicate that incorporating refined domain knowledge enhances both the accuracy and robustness of the model. 

In particular, compared with the performance of models using three input features, the performance of AINN shows a trend of degradation. This may be because AINN cannot effectively distinguish soil features collected from different depths, which negatively affects its ability to predict N$_2$O emissions. In contrast, KAINN processes input features from different soil depths separately, which helps preserve depth-specific information and leads to improved performance. However, since most bioreactions occur in soil layers around 10 cm, the overall improvement remains moderate.

\subsubsection{Performance Analysis with Ninth Features}

\begin{table}[htbp]
	\centering
	\small
	\setlength{\tabcolsep}{3pt}
	\caption{Performance comparison of various models across validation (2022) and testing (2023) datasets with 9 features. Best results within each (year, step) group are highlighted in bold.}
	\begin{tabular}{l c c c c | c c c}
		\toprule
		& & \multicolumn{3}{c}{2022} & \multicolumn{3}{c}{2023} \\
		\cmidrule(lr){3-5} \cmidrule(lr){6-8}
		Variant & Step & RMSE & MAE & R$^2$ & RMSE & MAE & R$^2$ \\
		\midrule
		CNN & 12 & $38.163 \pm 5.757$ & $26.625 \pm 5.782$ & $0.468 \pm 0.164$ & $22.766 \pm 4.794$ & $18.462 \pm 4.858$ & $-0.797 \pm 0.763$ \\
		CNN-DLEM & 12 & $35.493 \pm 12.057$ & $22.312 \pm 8.099$ & $0.502 \pm 0.435$ & $11.895 \pm 2.835$ & $7.280 \pm 1.659$ & $0.504 \pm 0.266$ \\
		CNN-KDLEM & 12 & $\mathbf{28.378 \pm 0.367}$ & $\mathbf{16.739 \pm 0.400}$ & $\mathbf{0.712 \pm 0.007}$ & $\mathbf{10.609 \pm 0.345}$ & $\mathbf{5.005 \pm 0.597}$ & $\mathbf{0.624 \pm 0.025}$ \\
		\midrule
		CNN & 24 & $38.214 \pm 6.833$ & $26.419 \pm 6.504$ & $0.462 \pm 0.205$ & $22.266 \pm 5.467$ & $17.909 \pm 5.473$ & $-0.742 \pm 0.884$ \\
		CNN-DLEM & 24 & $32.150 \pm 1.307$ & $19.658 \pm 0.939$ & $0.630 \pm 0.030$ & $11.091 \pm 1.060$ & $7.411 \pm 1.242$ & $0.587 \pm 0.079$ \\
		CNN-KDLEM & 24 & $\mathbf{28.256 \pm 0.578}$ & $\mathbf{16.569 \pm 0.508}$ & $\mathbf{0.714 \pm 0.012}$ & $\mathbf{10.616 \pm 0.330}$ & $\mathbf{4.806 \pm 0.520}$ & $\mathbf{0.624 \pm 0.024}$ \\
		\midrule
		CNN & 48 & $37.289 \pm 4.340$ & $25.915 \pm 4.350$ & $0.496 \pm 0.120$ & $21.562 \pm 4.106$ & $17.276 \pm 4.179$ & $-0.600 \pm 0.591$ \\
		CNN-DLEM & 48 & $32.532 \pm 1.551$ & $20.137 \pm 0.847$ & $0.620 \pm 0.036$ & $11.057 \pm 2.023$ & $6.900 \pm 2.243$ & $0.580 \pm 0.167$ \\
		CNN-KDLEM & 48 & $\mathbf{28.656 \pm 0.339}$ & $\mathbf{16.949 \pm 0.372}$ & $\mathbf{0.706 \pm 0.007}$ & $\mathbf{10.510 \pm 0.316}$ & $\mathbf{4.900 \pm 0.586}$ & $\mathbf{0.632 \pm 0.022}$ \\
		\midrule
		\midrule
		LSTM & 12 & $37.678 \pm 6.295$ & $25.983 \pm 5.723$ & $0.479 \pm 0.184$ & $20.296 \pm 6.103$ & $16.080 \pm 6.008$ & $-0.485 \pm 0.902$ \\
		LSTM-DLEM & 12 & $35.590 \pm 11.999$ & $22.118 \pm 8.253$ & $0.500 \pm 0.434$ & $12.137 \pm 2.631$ & $6.997 \pm 1.932$ & $0.488 \pm 0.254$ \\
		LSTM-KDLEM & 12 & $\mathbf{27.750 \pm 0.529}$ & $\mathbf{16.203 \pm 0.376}$ & $\mathbf{0.724 \pm 0.011}$ & $\mathbf{10.877 \pm 0.391}$ & $\mathbf{4.825 \pm 0.397}$ & $\mathbf{0.605 \pm 0.029}$ \\
		\midrule
		LSTM & 24 & $38.998 \pm 5.577$ & $26.914 \pm 4.538$ & $0.446 \pm 0.160$ & $21.360 \pm 5.644$ & $17.217 \pm 5.476$ & $-0.616 \pm 0.829$ \\
		LSTM-DLEM & 24 & $29.843 \pm 0.957$ & $19.000 \pm 0.793$ & $0.681 \pm 0.020$ & $12.007 \pm 1.136$ & $7.535 \pm 1.015$ & $0.516 \pm 0.092$ \\
		LSTM-KDLEM & 24 & $\mathbf{27.755 \pm 0.601}$ & $\mathbf{16.816 \pm 0.533}$ & $\mathbf{0.724 \pm 0.012}$ & $\mathbf{10.894 \pm 0.390}$ & $\mathbf{5.592 \pm 0.801}$ & $\mathbf{0.604 \pm 0.029}$ \\
		\midrule
		LSTM & 48 & $36.454 \pm 3.345$ & $24.700 \pm 2.487$ & $0.521 \pm 0.087$ & $19.457 \pm 3.390$ & $15.522 \pm 3.085$ & $-0.296 \pm 0.446$ \\
		LSTM-DLEM & 48 & $29.978 \pm 0.777$ & $18.772 \pm 0.523$ & $0.678 \pm 0.017$ & $12.845 \pm 1.655$ & $7.882 \pm 1.146$ & $0.442 \pm 0.149$ \\
		LSTM-KDLEM & 48 & $\mathbf{26.955 \pm 0.564}$ & $\mathbf{16.558 \pm 0.800}$ & $\mathbf{0.740 \pm 0.011}$ & $\mathbf{10.793 \pm 0.513}$ & $\mathbf{5.776 \pm 0.850}$ & $\mathbf{0.611 \pm 0.038}$ \\
		\midrule
		\midrule
		Transformer & 12 & $33.254 \pm 1.922$ & $23.278 \pm 2.117$ & $0.603 \pm 0.046$ & $17.286 \pm 3.040$ & $14.195 \pm 2.635$ & $-0.024 \pm 0.362$ \\
		Transformer-DLEM & 12 & $30.689 \pm 1.717$ & $19.738 \pm 1.915$ & $0.662 \pm 0.037$ & $12.363 \pm 1.281$ & $8.765 \pm 2.139$ & $0.486 \pm 0.106$ \\
		Transformer-KDLEM & 12 & $\mathbf{27.469 \pm 0.757}$ & $\mathbf{16.924 \pm 0.541}$ & $\mathbf{0.730 \pm 0.015}$ & $\mathbf{11.943 \pm 0.438}$ & $\mathbf{7.657 \pm 0.656}$ & $\mathbf{0.524 \pm 0.035}$ \\
		\midrule
		Transformer & 24 & $35.526 \pm 6.108$ & $25.770 \pm 5.361$ & $0.536 \pm 0.171$ & $19.777 \pm 5.461$ & $16.695 \pm 5.165$ & $-0.393 \pm 0.826$ \\
		Transformer-DLEM & 24 & $29.804 \pm 1.182$ & $19.539 \pm 1.360$ & $0.682 \pm 0.025$ & $12.211 \pm 0.740$ & $8.831 \pm 1.236$ & $0.501 \pm 0.059$ \\
		Transformer-KDLEM & 24 & $\mathbf{26.738 \pm 1.081}$ & $\mathbf{16.927 \pm 0.768}$ & $\mathbf{0.744 \pm 0.020}$ & $\mathbf{12.180 \pm 0.527}$ & $\mathbf{7.689 \pm 0.542}$ & $\mathbf{0.505 \pm 0.044}$ \\
		\midrule
		Transformer & 48 & $32.996 \pm 3.125$ & $23.730 \pm 3.092$ & $0.607 \pm 0.080$ & $16.386 \pm 2.650$ & $13.532 \pm 2.480$ & $0.084 \pm 0.326$ \\
		Transformer-DLEM & 48 & $28.781 \pm 0.786$ & $18.226 \pm 0.934$ & $0.703 \pm 0.016$ & $\mathbf{12.046 \pm 0.722}$ & $\mathbf{8.225 \pm 1.197}$ & $\mathbf{0.515 \pm 0.059}$ \\
		Transformer-KDLEM & 48 & $\mathbf{27.114 \pm 0.996}$ & $\mathbf{16.898 \pm 0.835}$ & $\mathbf{0.737 \pm 0.020}$ & $13.132 \pm 1.471$ & $8.681 \pm 0.991$ & $0.419 \pm 0.143$ \\
		\midrule
		\bottomrule
	\end{tabular}
	\label{tab:ainn_9_features_block}
\end{table}

As shown in Table~\ref{tab:ainn_9_features_block}, the performance of CNN-, LSTM-, and Transformer-based models using nine input features is presented across the validation (2022) and testing (2023) datasets.

Overall, the observations with nine features are largely consistent with those obtained from the three- and six-feature settings. KAINN continues to demonstrate improved predictive performance over AINN in CNN- and LSTM-based architectures, while purely data-driven models remain less competitive.

However, compared with the previous settings, several differences can be observed. First, the overall performance of AINN degrades further, particularly for CNN- and LSTM-based models. In contrast, this type of degradation is less evident in Transformer-based models. One possible reason is that AINN does not treat features from different soil depths separately, which leads to larger prediction errors.

Meanwhile, the overall performance of KAINN shows slight improvement, although the gain is not significant. This may be because most bioreactions do not occur at deeper soil layers (e.g., around 30 cm), and thus the additional features contribute limited new information. However, an important observation is that the inclusion of additional features does not degrade the performance of KAINN. This indicates that KAINN is more resistant to overfitting than AINN.

\subsection{Ablation Study Analysis}

To investigate the contribution of each mechanistic component in the proposed LSTM-KDLEM model, we conduct a comprehensive ablation study on the 2023 growing season dataset, using a sequence length of 36 time steps and three input features. In this setting, both KDLEM and ODLEM share the same LSTM backbone, ensuring that performance differences arise solely from the incorporation of mechanistic knowledge.

The ablation study consists of two complementary parts: (1) removal experiments, where individual components are excluded from the full model, and (2) addition experiments, where each component is used independently. The results are summarized in Table~\ref{tab:lstm_kdlem_ablation_2023_rmse_mae_r2}.

\begin{table}[htbp]
	\centering
	\begin{threeparttable}
		\caption{Ablation study of LSTM-KDLEM on the 2023 dataset. \\Best results are highlighted in bold.}
		\label{tab:lstm_kdlem_ablation_2023_rmse_mae_r2}
		\begin{tabular}{lccc}
			\toprule
			Variant & RMSE & MAE & R$^2$ \\
			\midrule
			LSTM-KDLEM
			& $\mathbf{10.527 \pm 0.250}$ 
			& $\mathbf{5.184 \pm 0.525}$ 
			& $\mathbf{0.630 \pm 0.018}$ \\
			
			no\_respiration\tnote{a} 
			& $10.640 \pm 0.433$ 
			& $5.361 \pm 0.687$ 
			& $0.622 \pm 0.031$  \\
			
			no\_diffusion
			& $10.601 \pm 0.337$ 
			& $5.541 \pm 0.631$ 
			& $0.625 \pm 0.024$  \\
			
			no\_temperature
			& $10.808 \pm 0.381$ 
			& $5.919 \pm 0.852$ 
			& $0.610 \pm 0.028$  \\
			
			no\_wfp
			& $10.797 \pm 0.817$ 
			& $5.891 \pm 1.232$ 
			& $0.609 \pm 0.061$  \\
			
			\midrule
			LSTM-DLEM
			& $11.035 \pm 0.574$ 
			& $6.418 \pm 0.921$ 
			& $0.593 \pm 0.043$  \\
			
			respiration\tnote{b}
			& $10.985 \pm 0.709$ 
			& $6.477 \pm 1.070$ 
			& $0.596 \pm 0.053$  \\
			
			diffusion
			& $10.611 \pm 0.264$ 
			& $5.788 \pm 0.644$ 
			& $0.624 \pm 0.019$ \\
			
			temperature 
			& $10.692 \pm 0.483$ 
			& $6.035 \pm 0.960$ 
			& $0.618 \pm 0.034$ \\
			
			wfp 
			& $10.635 \pm 0.372$ 
			& $5.408 \pm 0.782$ 
			& $0.623 \pm 0.027$ \\
			\bottomrule
		\end{tabular}
		
		\begin{tablenotes}
		\footnotesize
		\item[a] Variants prefixed with ``no\_'' indicate that the corresponding mechanism is removed from the full LSTM-KDLEM model.
		\item[b] Variants listed under LSTM-DLEM indicate that the corresponding mechanism is incorporated into the base LSTM-DLEM model.
		\end{tablenotes}
	\end{threeparttable}
\end{table}

The full LSTM-KDLEM achieves the best overall predictive performance, with the lowest RMSE (10.527), lowest MAE (5.184), and highest $R^2$ (0.630) among all variants. This confirms that integrating multiple mechanistic components significantly enhances the model’s ability to capture the underlying dynamics of soil emissions. Although the lowest ACF error (0.142) is achieved when using the WFP-only variant, but the ACF of LSTM-KDLEM is not so significant different from this one, it is possibly due to the stochastic effect. Therefore, we still considered that the full LSTM-KDLEM acheive the best possible performance across all the variants.

Removing individual components consistently degrades model performance, demonstrating that each mechanism contributes meaningful information. Among all components, temperature and WFP appear to be the most critical:

\begin{enumerate}
	\item Removing temperature leads to the largest degradation in accuracy, increasing RMSE to 10.808 and reducing $R^2$ to 0.610, indicating the importance of temperature-driven microbial activity in soil processes.
	\item Removing WFP also results in significant deterioration (RMSE = 10.797, $R^2$ = 0.609), highlighting the essential role of soil moisture in regulating emissions. This is consistent with the observation of the agriculture experts in the field, since most the bio-reactions are dependent on the WFP.
\end{enumerate}

However, removing the respiration component leads to a relatively smaller degradation (RMSE = 10.640, $R^2$ = 0.622), indicating that while respiration provides useful environmental information, its contribution is less dominant compared to moisture and temperature-related processes. Furthermore, the removal of the diffusion component causes a moderate but noticeable decline in performance (RMSE = 10.601, MAE = 5.541), suggesting that nutrient transport processes contribute to capturing spatial and temporal variations.

The addition ablation further highlights the importance of combining multiple mechanisms. The LSTM-DLEM baseline (no mechanistic enhancement) performs the worst overall (RMSE = 11.035, $R^2$ = 0.593), confirming that original LSTM-DLEM without adding domain knowledge to regulate each paramters, are insufficient for capturing complex soil dynamics under limited feature settings. Among single-component variants, diffusion, WFP,  achieve relatively better performance (RMSE $\approx$ 10.61–10.63, $R^2$ $\approx$ 0.623–0.624), indicating that transport and moisture processes are strong individual predictors. The temperature-only and respiration-only variants show weaker performance, suggesting that these processes alone cannot adequately explain emission dynamics.

Overall, the ablation study indicates that no single component can fully capture the complexity of soil emissions. The full model consistently outperforms all ablated variants, demonstrating that the integration of multiple mechanisms provides complementary information.

In addition, an interesting deviation can be observed in the Transformer-based models. Unlike the CNN and LSTM architectures, where KAINN consistently improves performance over AINN, the Transformer-based KDLEM does not always yield better results on the 2023 dataset. One possible explanation is that the Transformer architecture, due to its strong representation capacity and sensitivity to input scaling, may already capture certain temporal dependencies effectively, reducing the relative benefit of additional knowledge constraints. As a result, the integration of domain knowledge through KAINN may introduce additional modeling bias that does not align optimally with the learned attention patterns.

\subsection{Interpretability Analysis for AINN and KAINN}

\begin{figure}[htbp]
	\centering
	\subfigure[Interface Value of CNN DLEM model]{
		\includegraphics[width=0.9\textwidth]{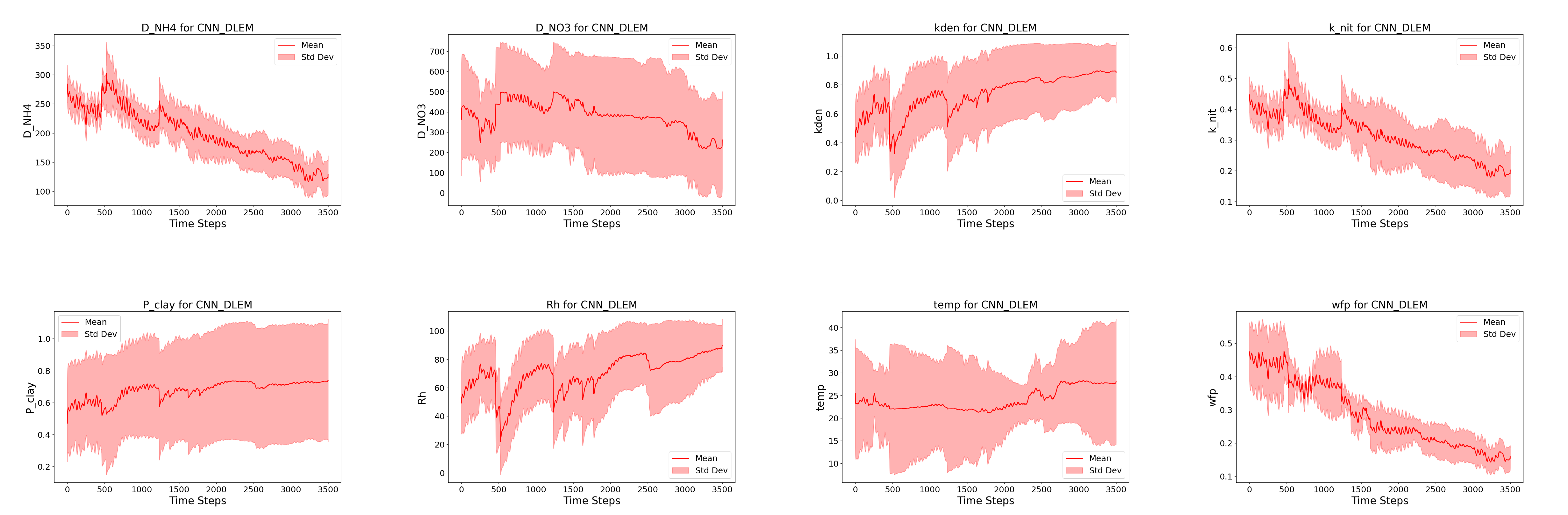}
		\label{fig:cnn_dlem_8}
	}
	\subfigure[Interface Value of CNN KDLEM model]{
		\includegraphics[width=0.9\textwidth]{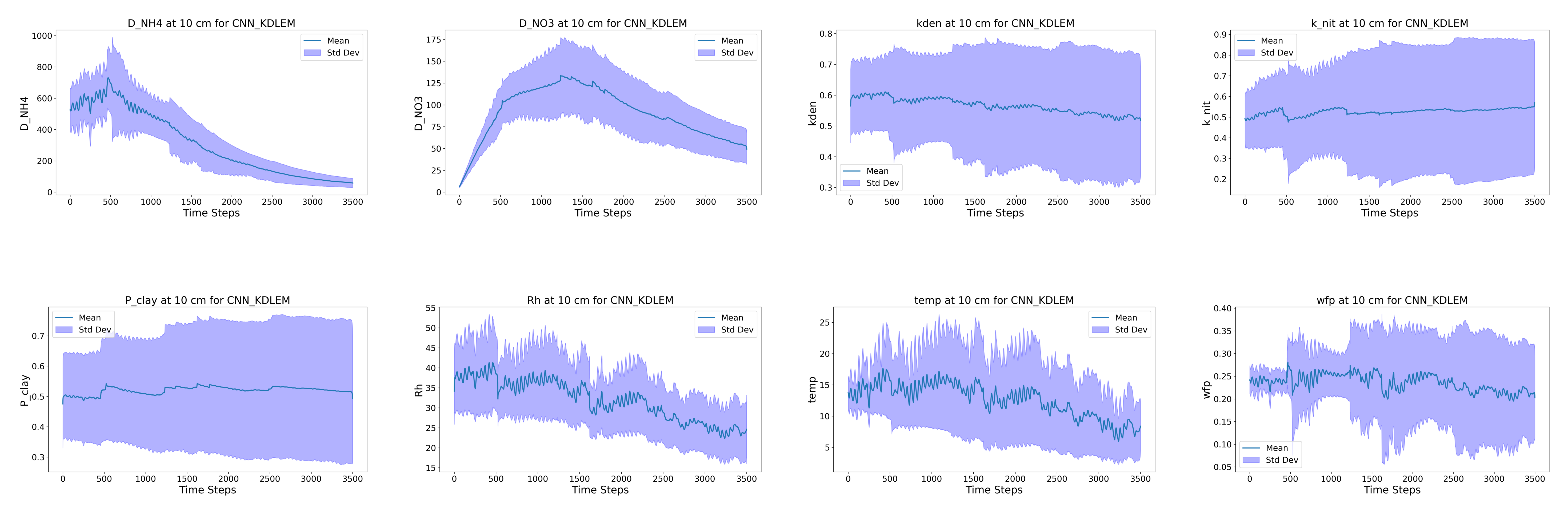}
		\label{fig:cnn_kdlem_8}
	}
	\caption{Interface value evolution of CNN-based models. CNN-KDLEM demonstrates reduced fluctuations compared with CNN-DLEM, indicating a constrained solution space with improved stability and interpretability. The evolution of NH$_4^+$ and NO$_3^-$ concentrations aligns with nitrification and denitrification processes.}
	\label{fig:interface_value_cnn_based}
\end{figure}

\begin{figure}[htbp]
	\centering
	\subfigure[Interface Value of Transformer DLEM model]{
		\includegraphics[width=0.9\textwidth]{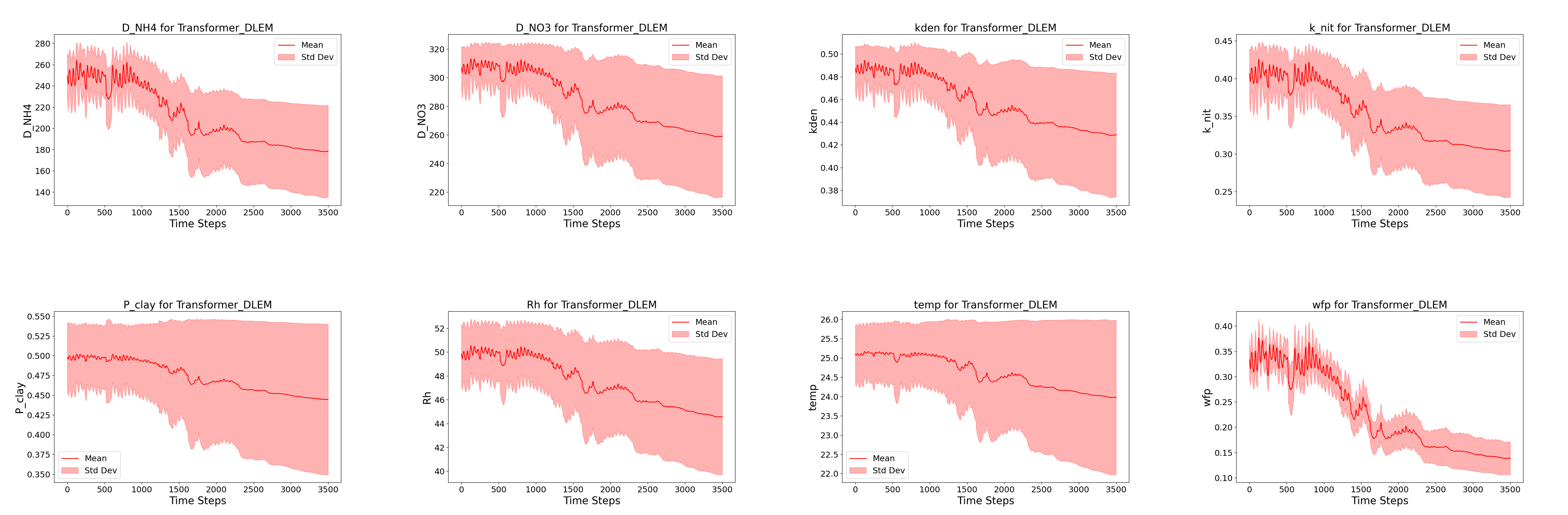}
		\label{fig:transformer_dlem_48}
	}
	\subfigure[Interface Value of Transformer KDLEM model]{
		\includegraphics[width=0.9\textwidth]{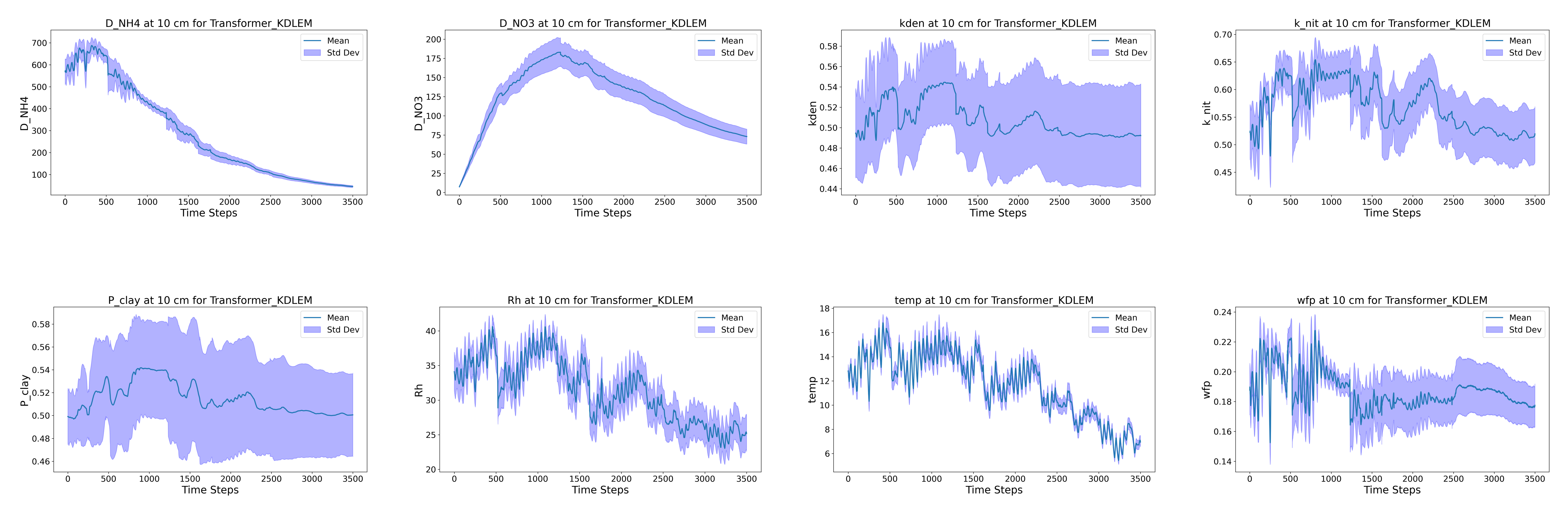}
		\label{fig:transformer_kdlem_48}
	}
	\caption{Interface value evolution of Transformer-based models. Transformer-KDLEM exhibits reduced fluctuations compared with Transformer-DLEM, indicating that domain knowledge integration constrains the solution space and improves the stability and interpretability of the interface variables. Compared with CNN-based models, Transformer-KDLEM captures these dynamics more accurately, providing deeper insight into nitrification and denitrification processes in soil systems.}
	\label{fig:interface_value_transformer_based}
\end{figure}

To further understand the effect of domain knowledge integration, we analyze the temporal evolution of the interface variables for CNN-DLEM, CNN-KDLEM, Transformer-DLEM, and Transformer-KDLEM, as illustrated in Figures~\ref{fig:interface_value_cnn_based} and \ref{fig:interface_value_transformer_based}. A clear and consistent pattern can be observed across both CNN- and Transformer-based architectures: the interface variables produced by KAINN exhibit significantly smoother trajectories with reduced fluctuations compared to those generated by AINN. In particular, the standard deviation bands associated with KAINN are generally narrower, indicating lower uncertainty in the estimated physical parameters. 

From a modeling perspective, this behavior can be attributed to the incorporation of domain knowledge in KAINN. By embedding physically meaningful constraints and relationships into the learning process, KAINN effectively restricts the solution space of the neural network. As a result, the model is guided toward physically plausible regions, reducing the variability of the inferred parameters and preventing unrealistic oscillations over time. In contrast, AINN relies more heavily on latent representations learned by the neural network, which are more sensitive to noise and data variability. This leads to larger fluctuations in the estimated parameters, as well as wider uncertainty bands. Such behavior suggests that the model explores a broader hypothesis space, which may improve flexibility but reduces interpretability and stability.

Another important observation is that KAINN produces more consistent temporal trends across different variables, such as $D_{\mathrm{NH4}}$ and $D_{\mathrm{NO3}}$. Since the fertilizer used is $\ce{CO(NH_2)_2}$, it gradually decomposes into $\ce{NH_4^{+}}$, which is then converted into $\ce{NO_3^{-}}$ through nitrification. At the same time, $\ce{NO_3^{-}}$ can be further transformed into \ce{N$_2$O} or absorbed by plants. Therefore, it is expected that the concentration of $\ce{NO_3^{-}}$ increases during the early stage and then decreases over time, as illustrated in Figures~\ref{fig:cnn_kdlem_8} and \ref{fig:transformer_kdlem_48}.

Overall, the interface evolution analysis demonstrates that KAINN not only improves predictive performance but also enhances the stability, physical consistency, and interpretability of the model by constraining the solution space and reducing uncertainty.

\section{Conclusion and Future Work}\label{sec:conclusion}

In this paper, KAINN is proposed for predicting soil N$_2$O emissions by integrating domain knowledge into a AINN framework. Building upon the original AINN, the proposed approach incorporates domain knowledge, including fertilizer diffusion dynamics, soil Rh modeling, and WFP estimation, to explicitly represent key environmental processes within the learning pipeline.

Extensive experiments across multiple architectures (CNN, LSTM, and Transformer) and growing seasons demonstrate that KAINN consistently improves predictive performance over both purely data-driven models and the original AINN framework. In particular, KAINN achieves lower prediction errors and higher $R^2$ values, while exhibiting enhanced robustness under different environmental conditions and farming practices between validation and testing datasets. These results highlight the effectiveness of domain knowledge integration in improving generalization in complex environmental systems.

Beyond predictive accuracy, KAINN also provides improved interpretability. The analysis of interface variable evolution shows that KAINN produces smoother and more physically consistent trajectories with reduced uncertainty compared to AINN. This behavior can be understood from an information-theoretic perspective: by embedding domain knowledge, KAINN reduces the entropy of the hypothesis space and constrains the model toward physically plausible solutions, thereby improving both stability and reliability.Moreover, the proposed framework demonstrates strong flexibility across different input configurations. While increasing the number of input features may introduce additional noise for purely data-driven models, KAINN effectively leverages this information without performance degradation, indicating its robustness against overfitting and its ability to utilize heterogeneous environmental signals.

Despite these promising results, several limitations remain. The current study is based on data from a limited number of field sites and growing seasons, and further validation across diverse climatic and soil conditions is necessary. The parameters used to estimate fertilizer diffusion rates, soil Rh, and WFP may not be optimal; therefore, additional efforts are needed to identify parameter settings that generalize well across different scenarios. In addition, future work may explore more advanced uncertainty quantification methods, such as separating epistemic and aleatoric uncertainty, and extending the framework to incorporate additional biochemical processes and multi-scale interactions.

\bibliographystyle{ACM-Reference-Format}
\bibliography{reference}

\end{document}